\documentclass[twoside,11pt]{article}

\usepackage{jair,theapa,rawfonts}

\usepackage{todonotes}
\usepackage{amsmath}
\usepackage{amsfonts}
\usepackage{booktabs}
\usepackage{multirow}
\usepackage{wrapfig}
\usepackage{subfigure} 
\usepackage{caption}
\usepackage{lipsum}
\usepackage{booktabs}       % professional-quality tables
\usepackage{amsfonts}       % blackboard math symbols
\usepackage{nicefrac}       % compact symbols for 1/2, etc.
\usepackage{xcolor}         % colors
\usepackage{enumitem}
\usepackage{graphicx}
\usepackage{colortbl}
\usepackage{amsmath}
\usepackage{amssymb}
\usepackage{pifont}
\usepackage{multirow}
\usepackage{listings}
\usepackage{tabulary}
\usepackage{pdflscape}
\usepackage{rotating}
\usepackage{tabularray}
\definecolor{delim}{RGB}{20,105,176}
\definecolor{numb}{RGB}{106, 109, 32}
\definecolor{string}{rgb}{0.64,0.08,0.08}

\usepackage{adjustbox}
\newcolumntype{R}[2]{%
    >{\adjustbox{angle=#1,lap=\width-(#2)}\bgroup}%
    l%
    <{\egroup}%
}
\newcommand{\guesswhat}{\textit{GuessWhat?!}}

\jairheading{1}{2023}{1-15}{6/91}{9/91}

\ShortHeadings{Visually Grounded Language Learning: a review of language games, datasets, tasks, and models}{A.Suglia and I.Konstas and O.Lemon}
\firstpageno{1}

\begin{document}

\title{Visually Grounded Language Learning: a review of language games, datasets, tasks, and models}

\author{\name Alessandro Suglia \email a.suglia@hw.ac.uk \\
       \addr School of Mathematical and Computer Sciences \\
       Heriot-Watt University\\
       Edinburgh, Scotland (UK)
       \AND
       \name Ioannis Konstas \email i.konstas@hw.ac.uk \\
       \addr School of Mathematical and Computer Sciences \\
       Heriot-Watt University\\
       Edinburgh, Scotland (UK)
       \AND
       \name Oliver Lemon \email o.lemon@hw.ac.uk \\
       \addr School of Mathematical and Computer Sciences \\
       Heriot-Watt University\\
       Edinburgh, Scotland (UK)
       }

\maketitle

\begin{abstract}
In recent years, several machine learning models have been proposed. They are trained with a language modelling objective on large-scale text-only data. With such pretraining, they can achieve impressive results on many Natural Language Understanding and Generation tasks. However, many facets of meaning cannot be learned by ``listening to the radio" only. In the literature, many Vision+Language (V+L) tasks have been defined with the aim of creating models that can ground symbols in the visual modality. In this work, we provide a systematic literature review of several tasks and models proposed in the V+L field. We rely on Wittgenstein's idea of `language games' to categorise such tasks into 3 different families: 1) discriminative games, 2) generative games, and 3) interactive games. Our analysis of the literature provides evidence that future work should be focusing on interactive games where \textit{communication in Natural Language} is important to resolve ambiguities about object referents and action plans and that \textit{physical embodiment} is essential to understand the semantics of situations and events. Overall, these represent key requirements for developing grounded meanings in neural models. 
\end{abstract}

\section{Introduction}

Symbols of a language acquire meanings when used \emph{to do things in the world}~\cite{clark1996using}. In such cases, language is a cooperative enterprise used by humans to achieve specific goals. During such cooperative activities, humans coordinate meanings~\cite{clark1991grounding}. Meanings are therefore dynamic entities that humans agree upon in conversation. It is important to underline that learning a language cannot happen in isolation. Learning a language is achieved by humans engaged in activities such as \textit{language games}~\cite{wittgenstein1953philosophische}. Such language games involve one or more interlocutors, who use language as a communication protocol used to express preferences, goals, and execute \emph{actions}~\cite{austin1975things}. 

%In order to execute specific actions, the language game has to be \emph{embodied} in a given environment because the act of referring to something will depend upon the context in which this is completed. 

There have been many attempts at computationally representing the meanings of words in a language. From symbolic approaches (e.g., \cite{winograd1971procedures}), statistical approaches (e.g., \cite{landauer1997solution}) to distributed representations (e.g., \cite{elman1990finding}), AI researchers have managed to achieve some significant results.
%seemingly outstanding results. 
Particularly, after the recent introduction of large-scale neural language models (e.g., BERT~\cite{bert}, GPT~\cite{radford2019gpt2}), the goal of achieving real understanding seems closer than before. However, upon careful inspection and probing, such models demonstrate only a superficial level of understanding of Natural Language. Many researchers in the field of AI and, more specifically Computational Linguistics, have argued that real understanding cannot be achieved by exposing machines to text corpora only~\cite{bender2020climbing}. This would mean expecting somebody to learn a language by reading alone~\cite{egl}. They argue that meaning is a result of grounding symbolic representations via multimodal perceptual experiences of concepts in the world. Such a level of perceptual experience can be obtained only when the agent is \textit{embodied} in the environment. 

Motivated by the need to expose artificial agents to more sophisticated perceptual information, in this survey we focus on the visual modality as a source of perceptual information. In particular, we are interested in tasks that have been proposed to study the symbol grounding problem~\cite{harnad1990symbol} in situated and embodied visual contexts. The main contributions of this survey are the following:

\begin{enumerate}
\item We provide a novel categorisation of visually grounded language games based on the skills and capabilities required to solve them;
\item We apply the categorisation to a collection of ~50 datasets presented in the last 20 years in the literature on Artificial Intelligence and Natural Language Processing;
\item We report an analysis of ~51 recent visually grounded models that have been proposed to tackle the tasks studied in this work;
\item We propose relevant research questions that will guide future research in grounded language learning.

\end{enumerate}

Our survey is divided into several sections that are described as follows: Section~\ref{sec:background} presents a discussion of the background topics that are required to understand the importance of language games for V+L research; Section~\ref{sec:tasks_review} reports our task categorisation that analyses the V+L tasks presented so far in the literature; Section~\ref{sec:gll_models} presents a survey of the V+L models that are able to encode the vision and language modalities; finally Section~\ref{sec:discussion} presents a discussion on promising research directions for the future of V+L research; Section~\ref{sec:conclusions} concludes the paper.

\section{Visually Grounded Language Learning} \label{sec:background}

Teaching agents to understand Natural Language has been the main objective of several research projects since the early days of Artificial Intelligence~\cite{winograd1971procedures,newell1972human}. However, understanding the meanings of words in Natural Language can be brought back to the \textit{Symbol Grounding} problem \cite{harnad1990symbol}: 
\begin{quote}
    ``How can the meanings of the meaningless symbol tokens, manipulated solely on the basis of their (arbitrary) shapes, be grounded in anything but other meaningless symbols?"
\end{quote}

\begin{figure}[ht]
    \centering
    \includegraphics[scale=0.2, keepaspectratio]{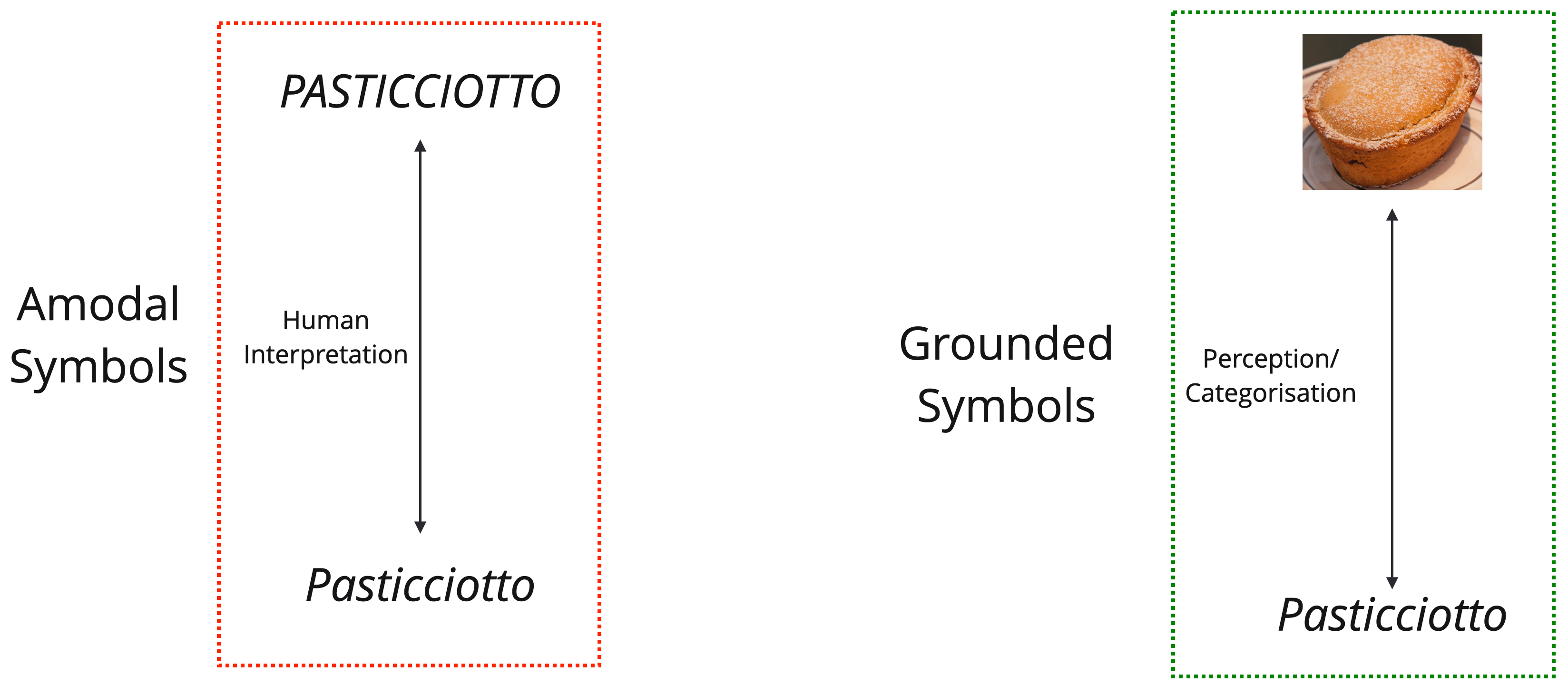}
    \caption{An example of the symbol grounding problem: a non-native speaker coming across the word ``pasticciotto" for the first time will struggle to understand it because it is expressed only via \textit{amodal symbols}. When the speaker receives an image associated with it, they are able to perform a \textit{perception/categorisation} step thanks to which the word is grounded in \emph{experience}, finally revealing its \textit{meaning}.}
    \label{fig:grounding_example}
\end{figure}

Associating a referent in the real world with a symbolic representation via a specific and contextualised meaning still represents a challenge for artificial agents. For instance in Figure~\ref{fig:grounding_example}, interpreting the meaning of the word ``pasticciotto" would not be possible without a \textit{categorisation} step that associates the meaningless symbols of the word to a \textit{concept} which stands for a composition of several perceptual features such as visual, functional, olfactory and gustatory features. These are a result of the agent's \textit{experience} of the concept in a given physical context.

This problem immediately implies that teaching computers Natural Language just by providing textual information might not be enough to effectively demonstrate that the agent \emph{understands} language just like humans do. Artificial agents should \textit{experience} the object whose symbols refer to, completing the so-called \textit{Semiotic Triad} defined by \cite{peirce1902logic} as follows: 
\begin{quote}
    ``I define a \textit{sign} as anything which is so determined by something else, called its \textit{Object}, and so determines an effect upon a person, which effect I call its \textit{interpretant}, that the latter is thereby immediately determined by the former." 
\end{quote}
From this statement we recognise the importance of the perceptual experience in the language learning process. The agent has to be exposed to a representation of the entity in the world (Object) that the word used (sign) refers to. The connection between the sign and the object becomes concrete once the object is perceived by the sensory organs of the agent. This process is what we consider as \textit{grounding}.

\begin{figure}
    \centering
    \includegraphics[scale=0.4]{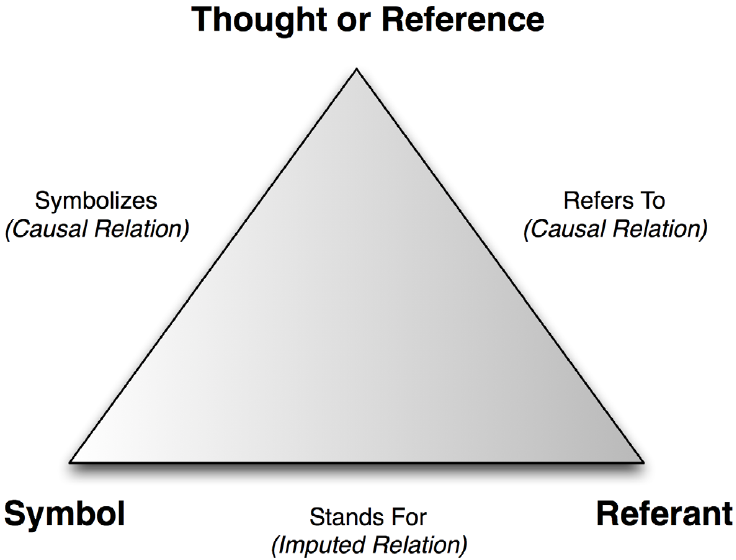}
    \caption{Semiotic Triad defined by \protect\cite{ogden1923meaning} which describes the relation that holds between symbols and objects in the world as well as the mental process that connects them.}
    \label{fig:semiotics_triad}
\end{figure}

The connection between language and thought was the main topic of interest of the seminal work by \cite{ogden1923meaning}. Particularly, they define \textit{Symbols} as the tools to organise, communicate and direct \textit{Thoughts}. As illustrated in the diagram in Figure~\ref{fig:semiotics_triad}, a symbol has a causal relationship with a thought that is instantiated when we speak. On the other hand, we use a symbol to refer to a given reference in the external world. Between the abstract \textit{Thought} and the \textit{Referent} (object in the world) a direct (or indirect) causal relation exists as well. 

\subsection{Compositionality \& Meaning} \label{sec:compositionality}

The term `compositionality' is defined as the following property of language: the ability to construct larger linguistic expressions by combining simpler parts. 
The focus of linguistic studies has been on \textit{semantic compositionality}, i.e., the principle whereby the meaning of a linguistic expression is a function of the meaning of its components and the rules used to combine them \cite{montague1970universal}. This ability to recombine known grounded symbols is what enables the grounding of complex expressions to occur. Following the example in \cite{harnad1990symbol}, suppose we are aware that both the symbols ``horse" and ``stripes" are grounded by appropriate representations learned from experience. Now we can consider a new category ``zebra" which is obtained as a combination of the elementary categories as follows: ``zebra" = ``horse" \& ``stripes". The answer by \cite{harnad1990symbol} to the question ``What is the representation of a zebra?" is as follows:

\begin{quote}
``It is just the symbol string `horse \& stripes'. But because `horse' and `stripes' are grounded, `zebra' inherits the grounding, through its grounded symbolic representation. In principle, someone who had never seen a zebra (but had seen and learned to identify horses and stripes) could identify a zebra on first acquaintance armed with this symbolic representation alone."
\end{quote}

This ability to combine grounded representations to generate other grounded representations is the result of associating meaning with a given representation. In this sense, compositionality can be considered as a more general learning bias that humans leverage in order to learn new tasks by reusing their knowledge about previously learned tasks and concepts \cite{fodor1988connectionism,fodor2002compositionality}. Indeed, compositionality represents a key element to cope with the huge amount of labelled data required by classical Deep Learning algorithms \cite{marcus2018deep} which learn tasks in an \textit{end-to-end} fashion, directly from the raw data~\cite{lecun2015:deep_learning}. Therefore, we are interested in learning representations of concepts that are \textit{grounded in perceptual experience} and also demonstrate some \textit{degree of compositionality}.

A prerequisite for learning and developing meanings is to develop grounded representations that can favour systematic generalisation to novel combinations of concepts that we have never seen before. Possessing such flexibility assumes that such learned conceptual representations should, by definition, not be static. They must dynamically change depending on the context and signals that the agent receives when interacting with the environment. Compositionality comes into play when an agent has to dynamically refine its internal representations when tasked to coordinate with another agent to collaboratively achieve a goal~\cite{lake2021word}. In such Natural Language interactions, words acquire meanings that are instantiated in the ways in which they are used as part of a systematic communication protocol which is \textit{a language}.

% % \todo[inline]{Importance of compositionality for language understanding}
% \todo[inline]{Importance of compositionality for dynamic representations that can be adapted and combined}
% \todo[inline]{Meaning is in their use, we need to learn to use words as part of a communication protocol which is a language}

\subsection{Language Games}

% Can be easily deleted for now
% In order to understand how to learn a language, we need to understand what is intended as ``language" first. The term ``language'' can be more generally defined as a system of communication, primarily used by humans, based on speech and gesture, sign, or often writing~\cite{good2013languoid}. Human language has been studied in many different fields due to its importance for the development of human intelligence. Philosophers as well as Linguists, considered the ability to use language as an innate ability~\cite{berwick2013evolution,chomsky2014aspects} which is wired in our brains when we are born. In contrast, in this work, we explore the hypothesis that ``human communication is thus a fundamentally cooperative enterprise, operating most naturally and smoothly within the context of (1) mutually assumed common conceptual ground, and (2) mutually assumed cooperative communicative motives"~\cite{tomasello2010origins}. 

The definition of language as a tool for human communication underlines the social functions of language. Additionally, humans use it to express themselves and to manipulate objects in their environment. Therefore, learning a language means being able to communicate effectively using a predefined communication protocol based on symbols and transformations of them. The meaning associated with each symbol can be acquired by engaging in goal-oriented conversations with other agents. This practice is what \cite{wittgenstein1953philosophische} defined as ``language games". Therefore, language functions within the active, practical lives of speakers involved, and is  deeply connected (i.e., \emph{grounded}) with non-linguistic elements and behaviours that are part of their environment. Indeed, \cite{wittgenstein1953philosophische} considers language as a system of signs that acquire meaning \emph{in situ}, embedded in speakers' lives.

When completing real-world tasks, artificial agents must \emph{communicate} in order to understand what the goal and requirements of a certain task are. Therefore, in this work, we underline the importance of the hypothesis that ``human communication is thus a fundamentally cooperative enterprise, operating most naturally and smoothly within the context of (1) mutually assumed common conceptual ground, and (2) mutually assumed cooperative communicative motives"~\cite{tomasello2010origins}. This definition highlights the importance of using the terminology of language games as an experimental framework for understanding the complexity, properties, and importance of certain tasks for the overall goal of language understanding and visual grounding. 

In contrast to fully symbolic approaches which argue that language can be defined as an abstract symbol manipulation system~\cite{newell1972human,pylyshyn1986computation,fodor1983modularity}, \cite{wittgenstein1953philosophische} considers language as a system that has a function, and acquires meaning, only when it is actually used to do things in the world. As in the \emph{Builder} example~\cite{wittgenstein1953philosophische}, the phrase ``pick up" will acquire meaning only when the \textit{Architect} will instruct the \textit{Builder} to pick a specific object up.

In this survey, we underline the importance of the functional and goal-oriented nature of Natural Language, investigating how artificial agents can acquire visually grounded representations \emph{by playing} situated and embodied language games.

\section{Visually Grounded Language Learning Tasks} \label{sec:tasks_review}

This paper surveys the state of the art in Visually Grounded Language Learning, which is a type of \textit{Interactive Grounded Language Learning}.
% After defining the intrinsic properties of language games and how they relate to the concept of grounding, we define the setup in which agents can acquire such grounded language. This setup focuses on the problem of \textit{Interactive Grounded Language Learning}. 
This can be considered as an instance of \textit{Interactive Task Learning} \cite{gluck2018interactive}, i.e., ``any process by which an \textit{agent (A)} improves its \textit{performance (P)} on some \textit{task (T)} through \textit{experience (E)}, when E consists of a series of sensing, effecting, and communication interactions between A, its world, and crucially other agents in the world''. We explore the idea of Wittgenstein's \textit{language games} \cite{wittgenstein1953philosophische} as a means to assess the linguistic capabilities of interactive learning agents. Specifically, as in the case of the ``Talking Heads'' experiments \cite{steels2015talking}, we explore the idea that linguistic capabilities could be assessed by 3 relevant language games: 1) \textit{guessing games} that require the agent to guess an unknown object in a scene; 2) \textit{action games} that involve the execution of actions by an agent as requested by a specific command; and 3) \textit{descriptive games} which require that the agent is able to describe the scene that it observes. 

We report an analysis of the state-of-the-art in terms of datasets and tasks that were proposed to study the problem of visual grounding for the English language (summarised in Table~\ref{tab:dataset_analysis}). We will divide the tasks into 3 macro categories following the different types of language games: 1) \textit{discriminative} tasks; 2) \textit{generative} tasks; and 3) \textit{interactive} tasks. Furthermore, we analyse each paper along the following key dimensions:
\begin{itemize}
    \item \textbf{Embodied}: is the agent able to explore, perceive and act in the environment?
    \item \textbf{Discriminative}: does the agent have to produce a single output by selecting it from a given set of candidates?
    \item \textbf{Generative}: does the agent have to produce a sequence of outputs conditioned on its input?
    \item \textbf{Interaction with environment}: is the agent able to manipulate and change the state of the environment?
    \item \textbf{Interaction with other agents}: does the agent have to communicate with other agents to solve the task?
\end{itemize}

%We now survey discriminative tasks, generative tasks, and interactive tasks.

% \usepackage{tabularray}
\begin{tiny}
\begin{longtblr}[
  caption = {Systematic analysis of several state-of-the-art datasets for Grounded Language Learning tasks. The analysis considers specific characteristics of the tasks and environments that are important for learning grounded meanings. We use ``x" to mark that a dataset satisfies a given property.},
  label = {tab:dataset_analysis}
]{
  width = \textwidth,
  colspec = {Q[200]Q[100]Q[150]Q[150]Q[150]Q[100]Q[100]Q[50]},
  cells = {c},
  vline{1,9} = {-}{},
  hline{1-2,67} = {-}{},
  rowhead=1
}

\textbf{Authors}                                  & \textbf{Dataset}    & \textbf{Embodied} & \textbf{Discriminative} & \textbf{Generative} & \textbf{Inter. with Env} & \textbf{Inter. with Agents} & \textbf{Size} \\
\cite{gordon2018iqa}             & IQA                 & yes               & x                       & x                   & x                        &                             & 75K           \\
\cite{Suhr2019:nlvr2}            & NLVR                & no                & x                       &                     &                          &                             & 100K          \\
\cite{suhr2019executing}         & CerealBar           & yes               &                         &                     & x                        & x                           & 1.2K          \\
\cite{alfred}                    & ALFRED              & yes               &                         & x                   & x                        &                             & 25K           \\
\cite{rxr}                       & RxR                 & yes               &                         &                     & x                        &                             & 126K          \\
\cite{padmakumar2021teach}       & TEACh               & yes               &                         &                     & x                        & x                           & 3.2K          \\
\cite{deitke2020robothor}        & RoboTHOR            & yes               &                         & x                   & x                        &                             &               \\
\cite{wortsman2019learning}      &                     & yes               &                         & x                   & x                        &                             &               \\
\cite{jain2019two}               &                     & yes               &                         & x                   & x                        &                             &               \\
\cite{kottur2021simmc}           & SIMCC 2.0           & no                & x                       &                     &                          & x                           & 11K           \\
\cite{chen2019touchdown}         & Touchdown           & yes               &                         & x                   &                          &                             & 9.3K          \\
\cite{haber2019photobook}        & Photobook           & no                &                         &                     &                          & x                           & 2.5K          \\
\cite{hermann2017grounded}       &                     & yes               &                         & x                   &                          &                             &               \\
\cite{yan2018chalet}             & CHALET              & yes               &                         &                     &                          &                             &               \\
\cite{mirowski2019streetlearn}   & StreetLearn         & yes               &                         &                     &                          &                             &               \\
\cite{de2018talk}                & Talk The Walk       & yes               &                         &                     & x                        & x                           & 10K           \\
\cite{kim2019codraw}             & CoDraw              & no                &                         &                     &                          & x                           & 10K           \\
\cite{brodeur2017home}           & HOME                & yes               &                         &                     &                          &                             &               \\
\cite{ramakrishnan2021habitat}   & HM3D                & yes               &                         &                     &                          &                             &               \\
\cite{savva2019habitat}          & Habitat             & yes               &                         &                     &                          &                             &               \\
\cite{kiela2020hateful}          & Hateful Memes       & no                & x                       &                     &                          &                             & 10K           \\
\cite{ruis2020benchmark}         & gSCAN               & no                &                         & x                   & x                        &                             & 300K          \\
\cite{shekhar2017foil}           & FOIL                & no                & x                       &                     &                          &                             & 297K          \\
\cite{zellers2019recognition}    & VCR                 & no                & x                       &                     &                          &                             & 290K          \\
\cite{park2020visualcomet}       & VisualCOMET         & no                & x                       &                     &                          &                             & 1.46M         \\
\cite{das2017visual}             & VisDial             & no                & x                       & x                   &                          &                             & 1.4M          \\
\cite{bogin2021covr}             & COVR                & no                & x                       &                     &                          &                             & 262K          \\
\cite{liu2021visually}           & MaRVL               & no                & x                       &                     &                          &                             & 5.6K          \\
\cite{narayan2019collaborative}  & Collab-Minecraft    & yes               &                         &                     & x                        & x                           & 509           \\
\cite{zarriess2016pentoref}      & PentoRef            & no                &                         &                     &                          & x                           & 1.3K          \\
\cite{pezzelle2019red}           & MALeViC             & no                & x                       &                     &                          &                             & 20K           \\
\cite{pezzelle2020different}     & BD2BB               & no                & x                       &                     &                          &                             & 18K           \\
\cite{huang2016visual}           &                     & no                &                         & x                   &                          &                             & 20.2K         \\
\cite{da2021edited}              & EMU                 & no                & x                       & x                   &                          &                             & 48K           \\
\cite{clark2021iconary}          & Iconary             & no                &                         &                     & x                        &                             & 55K           \\
\cite{goyal2017vqa2.0}           & VQA 2.0             & no                & x                       &                     &                          &                             & 1.1M          \\
\cite{hudson2019gqa}             & GQA                 & no                & x                       &                     &                          &                             & 22M           \\
\cite{kafle2017tdiuc}            & TDIUC               & no                & x                       &                     &                          &                             & 1.6M          \\
\cite{yang2021visual}            &                     & no                & x                       &                     &                          &                             & 53.2K         \\
\cite{kottur2019clevr}           & CLEVR-dialog        & no                &                         &                     &                          & x                           & 4.25M         \\
\cite{zellers2021piglet}         & PIGLeT              & yes               & x                       &                     &                          &                             & 280K          \\
\cite{abramson2020imitating}     & Playroom            & yes               &                         &                     & x                        & x                           &               \\
\cite{ilinykh2019tell}           & Tell-me-More        & no                &                         & x                   &                          &                             & 22K           \\
\cite{dobnik2020local}           & CUPS                & no                &                         &                     &                          & x                           & 1.3K          \\
\cite{yu2017burchak}             & BURCHAK             & no                &                         &                     &                          & x                           & 2.5K          \\
\cite{tokunaga2012rex}           & REX                 & no                &                         &                     & x                        & x                           & 9.8K          \\
\cite{chevalier2018babyai}       & BabyAI              & yes               &                         &                     & x                        &                             &               \\
\cite{cvdn}                      & CVDN                & yes               &                         &                     & x                        & x                           & 2K            \\
\cite{ilinykh2019meetup}         & MeetUp              & yes               &                         &                     & x                        & x                           & 430           \\
\cite{puig2020watch}             & Watch-And-Help      & yes               &                         &                     & x                        & x                           & 1.3K          \\
\cite{zheng2022spot}             & Spot the difference & no                &                         &                     & x                        &                             & 95K           \\
\cite{kiseleva2022interactive}   & IGLU                & yes               &                         &                     & x                        & x                           & 509           \\
\cite{bisk2016natural}           & BlocksWorld         & no                &                         & x                   & x                        &                             & 12K           \\
\cite{zhong2021silg}             & SILG                & no                &                         & x                   & x                        &                             &               \\
\cite{elliott2016multi30k}       & Multi30K            & no                &                         & x                   &                          &                             & 150K          \\
\cite{wang2021multisubs}         & Multisubs           & no                &                         & x                   &                          &                             & 4.5M          \\
\cite{kazemzadeh2014referitgame} & ReferitGame         & no                &                         & x                   &                          &                             & 20K           \\
\cite{johnson2016densecap}       & DenseCap            & no                &                         & x                   &                          &                             & 4.1M          \\
\cite{krishna2017dense}          & ActivityNet         & no                &                         & x                   &                          &                             & 20K           \\
\cite{thrush2022winoground}      & Winoground          & no                & x                       &                     &                          &                             & 1.6K          \\
\cite{bugliarello2022iglue}      & IGLUE               & no                & x                       &                     &                          &                             & 28K           \\
\cite{krojer2022image}           & ImageCoDe           & no                & x                       &                     &                          &                             & 22K           \\
\cite{Shridhar2021}              & ALFWorld            & yes               &                         & x                   & x                        &                             & 25K           \\
\cite{gao2022dialfred}           & Dialfred            & yes               &                         &                     & x                        & x                           & 53K           \\
\cite{gao2023alexa}              & Alexa Arena         & yes               &                         &                     & x                        & x                           & 46K           

\end{longtblr}

\end{tiny}

\subsection{Discriminative tasks} \label{sec:discriminative_tasks}

A discriminative task is defined as a language game in which the agent, given an image $I$ and a corresponding textual information $t$, has to select an option $\hat{c}$ among a set of candidates $\mathcal{C}$. The agent is successful if $\hat{c}$ matches the target option $c^*$. Such tasks have been the very first instance of grounded language learning tasks inspired by well-known image classification tasks in the Computer Vision community~\cite{deng2009imagenet}. 

\begin{figure}
    \centering
    \includegraphics[scale=0.5,keepaspectratio]{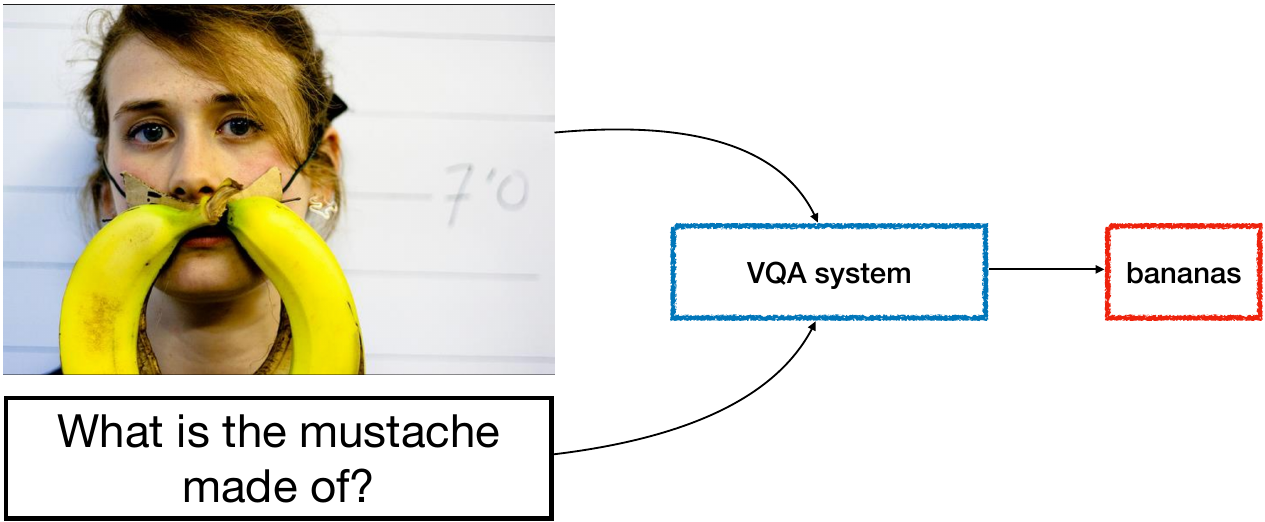}
    \caption{Example from the Visual Question Answering dataset \protect\cite{antol2015vqa}: the agent should learn to generate the answer ``bananas" given the image and the question.}
    \label{fig:vqa}
\end{figure}

%\subsubsection{Visual Question Answering (VQA)}
One of the most famous tasks in this category is {\bf Visual Question Answering (VQA)}. The generic formulation of this task, as shown in Figure~\ref{fig:vqa}, assumes that the agent receives a question about an image and has to provide an answer to it. The answer can be in free-form text. However, current implementations of VQA tasks, such as the one presented by \cite{antol2015vqa}, assume that the agent has a specific answer vocabulary (\textit{vocabulary-based VQA}) that it has to choose from. In this task, the agent should learn to reason about the objects in the image to answer the question. Questions usually require that the agent is aware of specific categories of objects, their colour, or their relative position in the image. However, the first release of the dataset had a bias according to which questions could be answered by relying only on one modality or only by relying on a specific subset of tokens in the question. For this reason, as described by \cite{goyal2017making}, a proper balancing procedure for the dataset is required. This prevents the case where high-capacity models obtain high accuracy only because they learn spurious correlations in the data. Other ways to mitigate this problem have been proposed. For instance, the TDIUC dataset~\cite{kafle2017tdiuc} divides the questions into 12 categories to have a better understanding of the model's capabilities. A related effort in this direction is the GQA dataset~\cite{hudson2019gqa}. GQA is a VQA dataset automatically generated by relying on Visual Genome scene graphs~\cite{krishna2017visual}. Having a structured representation of the scene adds a semantic layer on top of the purely perceptual one. Thanks to this additional level, it is possible to perform a more fine-grained analysis of the model's ability. For instance, the authors propose measures of \textit{consistency} of the model predictions (i.e., is the answer to semantically similar questions the same?), \textit{validity} (i.e., is the agent responding some colour to a colour question?), and \textit{plausibility} (i.e., does the answer make sense given the question?). However, because the problem is cast as a classification task, models learn to model the head of the answer probability distribution only. Therefore, they perform very poorly when generalising to out-of-distribution examples. To address this problem, specific datasets that test the generalisation abilities of VQA models have been proposed (e.g., \cite{agrawal2017cvqa}). The community proposed the CLEVR~\cite{johnson2017clevr} dataset to complete a more in-depth evaluation of the elementary visual reasoning skills of an agent using synthetic images generated using the Blender render engine.

\begin{figure}
    \centering
    \includegraphics[scale=0.35,keepaspectratio]{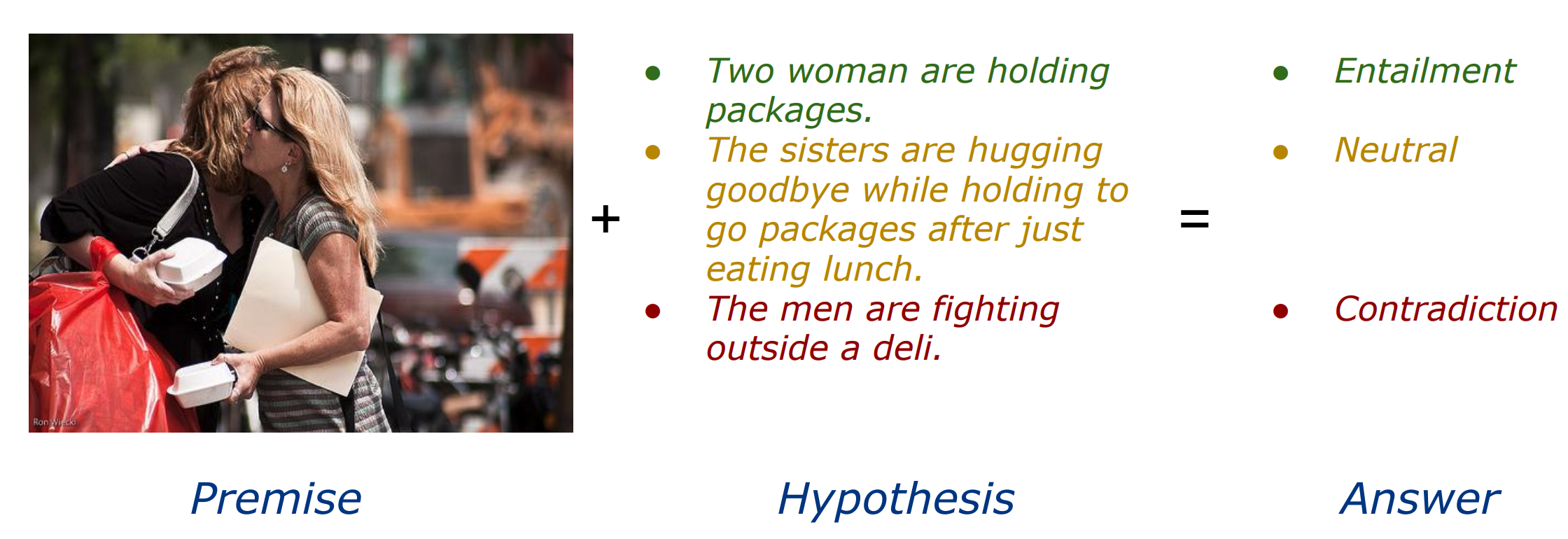}
    \caption{Example from the Visual Entailment task from \protect\cite{xie2018visual}. Given the state of the world (exemplified by the image), the agent has to verify, for each hypothesis, the validity of the statement. Figure adapted from \protect\cite{xie2018visual}.}
    \label{fig:vis_entailment}
\end{figure}

%\subsubsection{Visual Entailment (VE)}
Another instance of a discriminative task is {\bf Visual Entailment (VE)}~\cite{xie2018visual}. It is inspired by Textual Entailment tasks in Natural Language Processing (e.g., \cite{bowman2015large}) defined as follows: given a text premise $P$ and a text hypothesis $H$, the goal is to determine if $P$ implies $H$.  This task is usually cast as a 3-way classification problem where the label set includes ``entailment", ``neutral", or ``contradiction", based on the relation conveyed by the (P, H) text pair. As shown in Figure~\ref{fig:vis_entailment}, VE replaces premise $P$ with a real-world image. Based on this idea of entailment, several other tasks can be defined. For instance, the \textit{Cornell Natural
Language Visual Reasoning (NLVR)} dataset \cite{suhr2017corpus} is a corpus containing 92,244 sentence-image pairs, whose aim is to teach an agent whether a given statement is true or false for a given image. What makes this dataset interesting is that it includes several semantic phenomena such as cardinality (soft/hard) statements, existential and universal relations, as well as spatial relations. Sentences associated with the images are collected from real users, however, the corresponding images are synthetic and composed of simple coloured blocks. \cite{Suhr2019:nlvr2} extended this dataset with real-world images.

%\subsubsection{Find One mismatch between Image and Language caption (FOIL), and related tasks}

\begin{figure}
    \centering
    \includegraphics[scale=0.4, keepaspectratio]{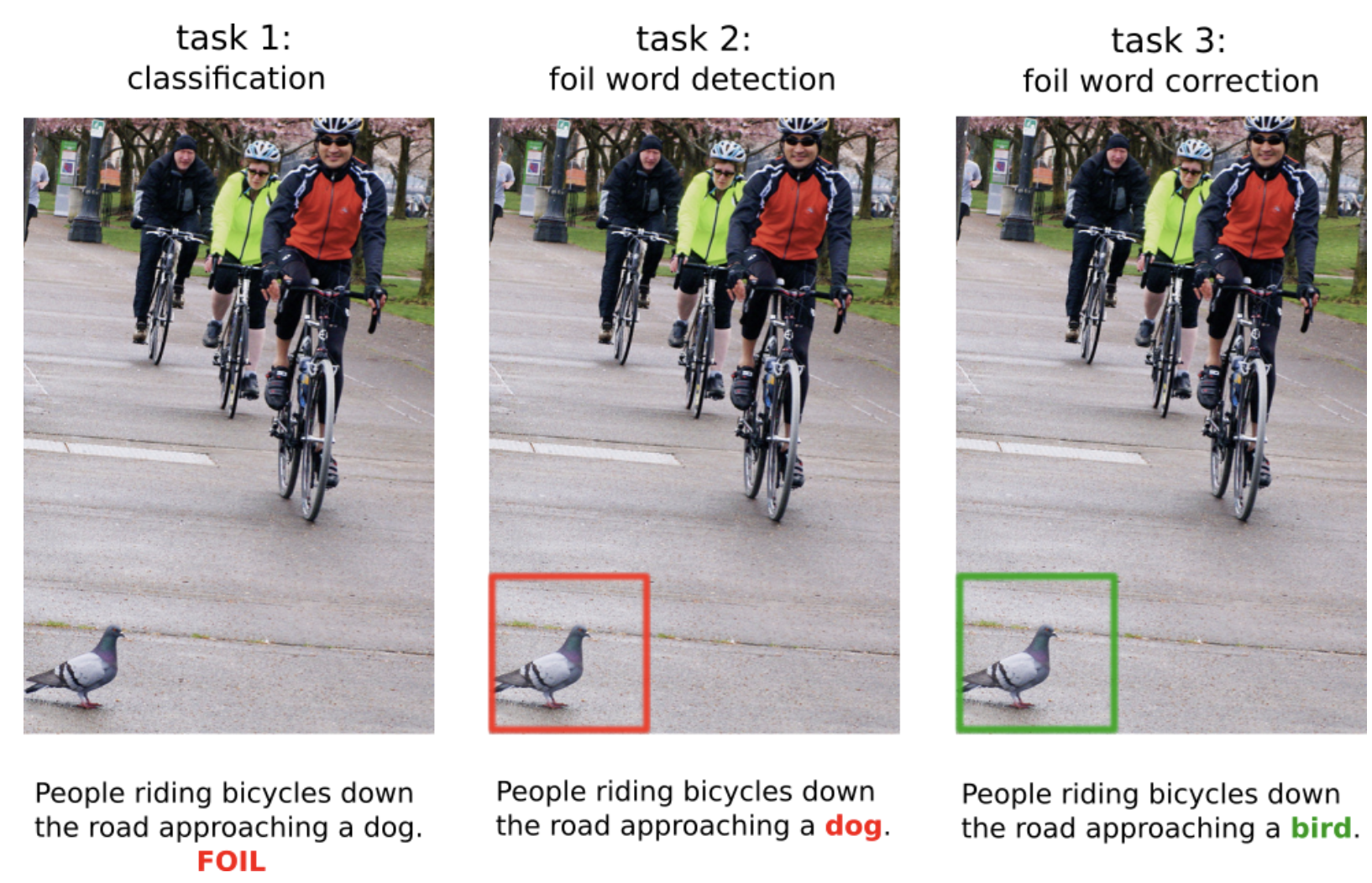}
    \caption{Example extracted from the FOIL dataset. It describes the 3 discriminative tasks: binary classification, foil word detection and foil word correction. Figure adapted from \protect\cite{shekhar2017foil}. }
    \label{fig:foil}
\end{figure}

Another discriminative task is the {\bf Find One mismatch between Image and Language caption (FOIL)} dataset \cite{shekhar2017foil}. It defines 3 different discriminative sub-tasks described as follows (see Figure~\ref{fig:foil} for an example): 

\begin{enumerate}
\item \textit{Binary Classification}: Given an image and a caption, the model is asked to mark whether the caption is correct, or wrong. The aim is to understand whether trained models can spot mismatches between their coarse representations of language and visual input;

\item \textit{Foil Word Detection}: Given an image and a caption, the model has to detect the foil word (i.e., a single word that is incorrect for the caption). The aim is to evaluate the understanding of the system at the word level;

\item \textit{Foil Word Correction}: Given an image, a caption and the foil word, the model has to detect the foil and provide its correction. The aim is to check whether the system's visual representation is fine-grained enough to be able to extract the information necessary to correct the error.
\end{enumerate}

\begin{figure}[ht]
    \centering
    \includegraphics[scale=0.275, keepaspectratio]{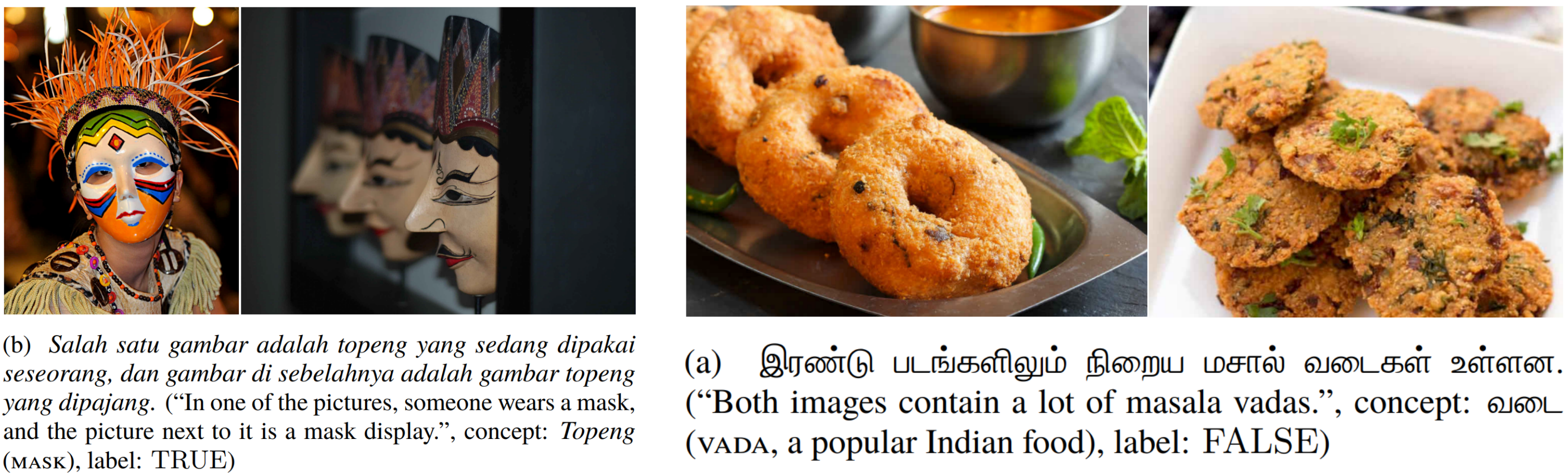}
    \caption{Examples extracted from the MaRVL dataset. The first image shows an example in Indonesian and the second in Tamil. Figures adapted from \protect\cite{liu2021visually}. }
    \label{fig:marvl}
\end{figure}

Most of these datasets have been annotated relying on image datasets that were derived from an English lexical database (i.e., WordNet~\cite{miller1995wordnet}). However, this imposes a bias on the data collection because all the elicited concepts will be familiar only to people with a European/American background. Since an over-arching goal in this area is to create general-purpose agents, they must rely on an unbiased source of training data. \cite{liu2021visually} proposed {\bf MaRVL}, a novel dataset for multicultural reasoning over vision and language. First of all, they propose a novel concept hierarchy that can be considered universal and not specific to the English language. After this step, they collect a dataset using a similar approach to \cite{Suhr2019:nlvr2}. The resulting dataset contains 5.6K captions using 5 typologically, geographically and genealogically diverse languages. As shown in Figure~\ref{fig:marvl}, this makes MaRVL a very challenging benchmark for multimodal and multilingual language understanding.

Similar to FOIL, {\bf MALeViC}~\cite{pezzelle2019red} defines a sentence verification task to study how Vision+Language (V+L) models can learn the meaning of gradable adjectives of size from different visual contexts. This benchmark includes several tasks that are specifically designed to study how an agent can learn to identify the reference set of a given statement in a specific visually-grounded context. This is a crucial skill that agents have to master when dealing with complex Embodied AI tasks involving a high level of ambiguity. Another task formulated as a discriminative language game is {\bf ``Be Different to Be Better"}~\cite{pezzelle2020different}. It is defined as a candidate selection task where the agent, based on a certain intention (i.e., their goal, attitude or feeling), has to choose, among a set of candidate actions, the one that a person would perform. Other instances of these tasks involve matching the correct caption to a given image. This is the case of \textbf{Winoground}~\cite{thrush2022winoground} where an agent has to be able to score higher the correct (image, caption) pair instead of the others inspired by the Winograd Schema Challenge~\cite{levesque2012winograd}. This dataset is extremely challenging because it contains images having an intricate visual structure that have been manually sourced by a team of experts. Humans are able to match the correct caption to its image while state-of-the-art models have a really hard time completing this task with top performance being just above random chance. 

\begin{figure}
    \centering
    \includegraphics[scale=0.5, keepaspectratio]{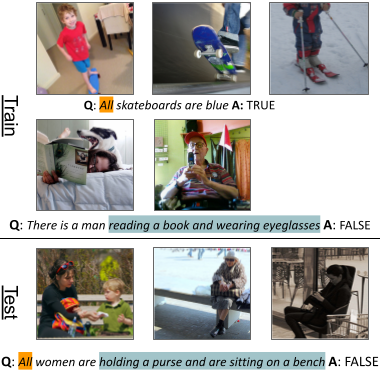}
    \caption{Example extracted from the COVR dataset. At training time, the agent is exposed to specific usages of quantifiers as well as object attributes. At test time, novel, unseen usages are presented so that the model has to \textit{systematically generalise} to novel statements. Figure adapted from \protect\cite{bogin2021covr}.}
    \label{fig:covr}
\end{figure}

The dataset \textbf{ImageCoDe} provides an additional benchmark for contextual image retrieval~\cite{krojer2022image} where the agent has to select the image that matches a contextual caption given a set of 9 distractors. As a result of their evaluation, they show that their benchmark is highly multi-modal (i.e., the agent requires both modalities to do well) and that current state-of-the-art V+L models are still far away from human-level performance. Most likely, this is because some of the situations require understanding the unfolding of events over time and cannot be learned from static images only. Similarly to the effort in VQA to inspect the visual grounding ability of the model, \cite{bogin2021covr} proposes a new benchmark for {\bf compositional language understanding (COVR)}. The task, inspired by NLVR~\cite{Suhr2019:nlvr2}, is formulated as a sentence verification problem having multiple reference images (see Figure~\ref{fig:covr}). The dataset has been created by defining specific splits having different compositional generalisation skills. It uses the GQA~\cite{hudson2019gqa} scene graphs to first find adversarial images to a target image. Then, it instantiates questions from language templates and finally uses manual validation and paraphrasing to generate the actual text. The generated examples require the agent to master the use of quantifiers and other operations, such as counting. This makes the task particularly challenging, especially in zero/few-shot splits where the agent has the chance to learn from a few examples of a given logical operator. 

\begin{figure}
    \centering
    \includegraphics[scale=0.21, keepaspectratio]{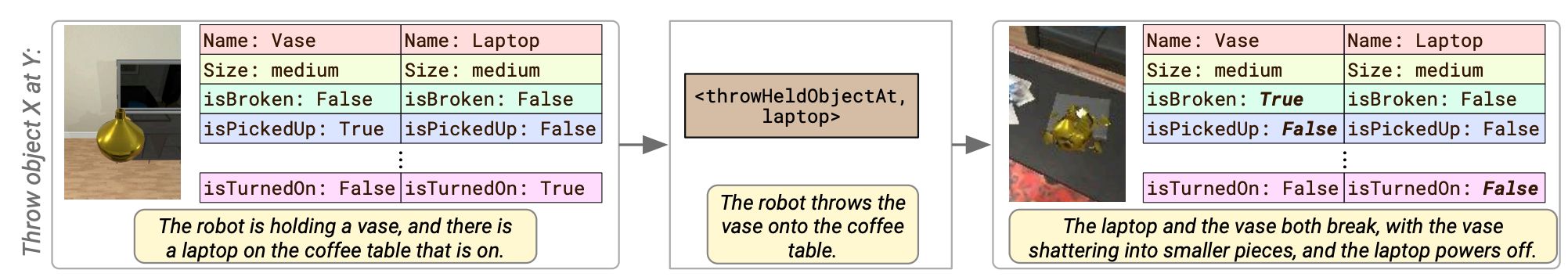}
    \caption{Example extracted from the PIGLeT dataset. Each example contains the representation of the state of the world before and after the action ``throw object" happens. The world state is expressed in terms of visual attributes derived from the 3D environment. Language annotations are available for each step of the transition. Figure adapted from \protect\cite{zellers2021piglet}.}
    \label{fig:piglet}
\end{figure}

Most of these tasks can be solved by relying on information immediately available in the image (e.g., the properties of the objects or the relationship between them). However, understanding the effects of given actions as well as the goal of specific actors in an image plays an important role in visual commonsense reasoning. To tackle this important problem, \cite{zellers2019recognition} propose the {\bf Visual Commonsense Reasoning} challenge. It is defined as a multiple-choice classification task where the agent has to select the correct answer to a given question as well as specify ``why" that answer was relevant. {\bf PIGLeT}~\cite{zellers2021piglet} studies the problem of learning physical commonsense knowledge. The authors automatically generated trajectories of actions using a planner and then annotated specific transitions where a state change happens. In particular, as shown in Figure~\ref{fig:piglet}, for each transition they collect a set of visual attributes for each state as well as a description of the action that triggered the state change, and a description of the final state. One of the main tasks involves predicting the attributes of the final state given the action executed by the agent. {\bf VisualCOMET}~\cite{park2020visualcomet} is another dataset for visual commonsense reasoning in static images. Given an image, a person that appears in it is selected and multiple reasoning inferences are annotated that are supposed to describe: 1) what needed to happen before; 2) the intents of the person; and 3) what will happen next.

Another multimodal dataset that requires higher-order reasoning skills is the {\bf Hateful Memes} challenge~\cite{kiela2020hateful}. This dataset was created to develop models that are able to detect internet abuse, an important problem in current social media platforms where sarcasm and other very subtle visual and language cues are used in an offensive manner. This problem is particularly important because it underlines the fact that words can assume different meanings depending on the multi-modal context in which they are used. Being able to derive their meaning is therefore a highly complex reasoning task.
%This problem is equally important because truly understanding how a word can assume a different meaning depending on a specific multi-modal context in which it is used.

\subsection{Generative tasks} \label{sec:generative_tasks}

% formal definition of a generative task (report the fact that is associated with descriptive games)
\textit{Generative} (or \textit{descriptive}) language games are formally defined by assuming that there is an agent that receives an image $I$ and has to generate a sequence of tokens $\hat{t} = \langle t_1, t_2, \dots, t_n \rangle$. The task is evaluated by considering a specific semantic similarity measure $S$ that determines how close $\hat{t}$ is to the ground-truth target sequence $t$. 

\begin{figure}[ht]
    \centering
    \includegraphics[scale=0.4,keepaspectratio]{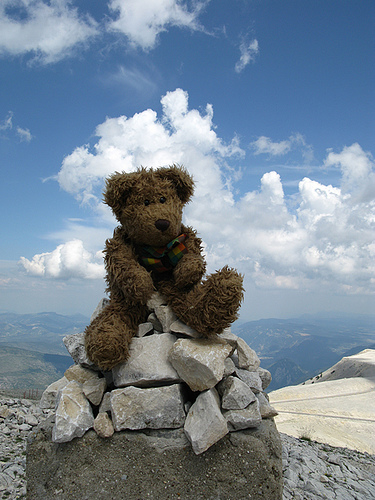}
    \caption{Image captioning example from the MSCOCO dataset \protect\cite{lin2014coco}. A possible caption for this image is: ``a teddy bear that has been placed on a pile of rocks."}
    \label{fig:image_captioning_example}
\end{figure}

One of the first tasks that have been proposed in the literature is {\bf Image captioning} in which the main objective is to generate a meaningful description of an input image (see Figure \ref{fig:image_captioning_example} for an example). One of the most used datasets for this task is MSCOCO \cite{lin2014coco} which provides 5 captions for every image as well as segmented objects. A similar dataset is \textbf{Flickr30K}~\cite{young2014image}. 
In a similar fashion, {\bf video captioning} tasks have been proposed (e.g., \cite{wang2019vatex,Lei2020tvr}) where either the subtitle or a caption has to be generated for a given video clip. A more complex scenario is represented by video summarisation (e.g., \cite{papalampidi2020screenplay}). Analogous to text summarisation~\cite{nenkova2011automatic}, video summarisation involves a more structured approach that resembles classic Natural Language Generation pipelines~\cite{reiter_dale_2000}. It usually involves several steps: 1) selecting the video frames that are worth considering for the summary; 2) aggregating the information in a coherent way to favour sound summaries; and 3) using the aggregated pieces of information to generate the text of an appropriate summary. Another task that involves multiple video frames as part of the input data is visual storytelling~\cite{huang2016visual}. In this case, the agent is provided with multiple images that describe the progression of some event over time. This task is more complex than video captioning because it is not only limited to factually describing the image but involves also adding some extra contextual cues (e.g., emotional and time-related aspects). Another generative task that involves describing a visual scene in multiple rounds is {\bf Tell-me-More}~\cite{ilinykh2019tell}. In their data collection, they ask their users to imagine they are talking to someone over the phone about an image they can see. This image belongs to a set of candidate images that are available to the other agent only. The receiver has to select which image the speaker is referring to. 

Most of the generation tasks described so far assume that the agent has to generate a description for the entire visual input. To evaluate how fine-grained visual representations are it is useful to focus only on certain parts of the image. This task is commonly called \textit{dense captioning} or \textit{referential expression} generation. In this task, an agent has to generate a description that applies to a very specific region of the image (i.e., a bounding box). For instance, \cite{kazemzadeh2014referitgame} propose the ReferIt game intended as a generic data collection procedure for referential expression generation. This procedure has been then applied to several datasets such as COCO~\cite{lin2014coco} and ImageCLEF~\cite{grubinger2006iapr}. Concurrently, rich semantic annotations of images have been collected via the VisualGenome project~\cite{krishna2017visual}. From these annotations, \cite{johnson2016densecap} released the \textbf{DenseCap} dataset to learn to generate descriptions associated with specific bounding boxes of the image. \cite{krishna2017dense} collected a dataset for captioning in videos as well. Particularly, this dataset was annotated by dividing the video into key activities (or moments) and then asking the annotator to report a sentence describing each activity. 

Another important research direction is to use visual information to boost performance in machine translation. The first dataset to be proposed for this task was \textbf{Multi30K}~\cite{elliott2016multi30k}. The dataset was created by creating German translations for each image description available in Flickr30K~\cite{young2014image}. However, this process was completed without providing images to the annotators introducing an apparent bias in the collected data~\cite{li2021vision}. To mitigate this issue, \cite{wang2021multisubs} proposed \textbf{Multisubs}, a large-scale multimodal and multilingual dataset that
facilitates research on grounding words to images in the context of their corresponding sentences. This dataset is based on movie subtitles but due to missing copyright, the authors didn't use the images of the video as reference. Instead, they relied on a multilingual knowledge base to find references to \textit{general} images about certain concepts (i.e., \textit{BabelNet}~\cite{navigli2010babelnet}).

\subsection{Interactive tasks}\label{sec:interactive_tasks}

In Section~\ref{sec:background}, we introduced the concept of language games as an experimental framework for studying machine learning tasks aimed at assessing certain language understanding skills of artificial agents. This is important because if we want to aim to create artificial agents that are trained to solve ecologically valid tasks~\cite{de2020towards}, we need to consider the fact that the real world contains: 1) sophisticated visual stimuli in the form of videos or images; 2) ambiguous, spontaneous, and incremental language that is typically used when communicating in natural language. 

The grounded language learning tasks that we described so far did not have a real connection with the concept of language games that we described in previous sections. Although each language game could be considered as \emph{situated} in the visual scene exemplified by the image, there was no notion of \emph{interaction} between the learning agent and the environment or other agents. Each task involved either generating one or multiple elements associated with the input. To get closer to the idea of language games described by \cite{wittgenstein1953philosophische}, we are interested in exploring \textit{interactive} tasks where the agent is either \textit{situated} or \textit{embodied} in the environment and has to interact in particular ways. This can be either by communicating with other agents to solve a given task or by executing actions in the environment to achieve a given goal. 

\subsubsection{Situated Interactive Games}

\begin{figure}
    \centering
    \includegraphics[scale=0.15,keepaspectratio]{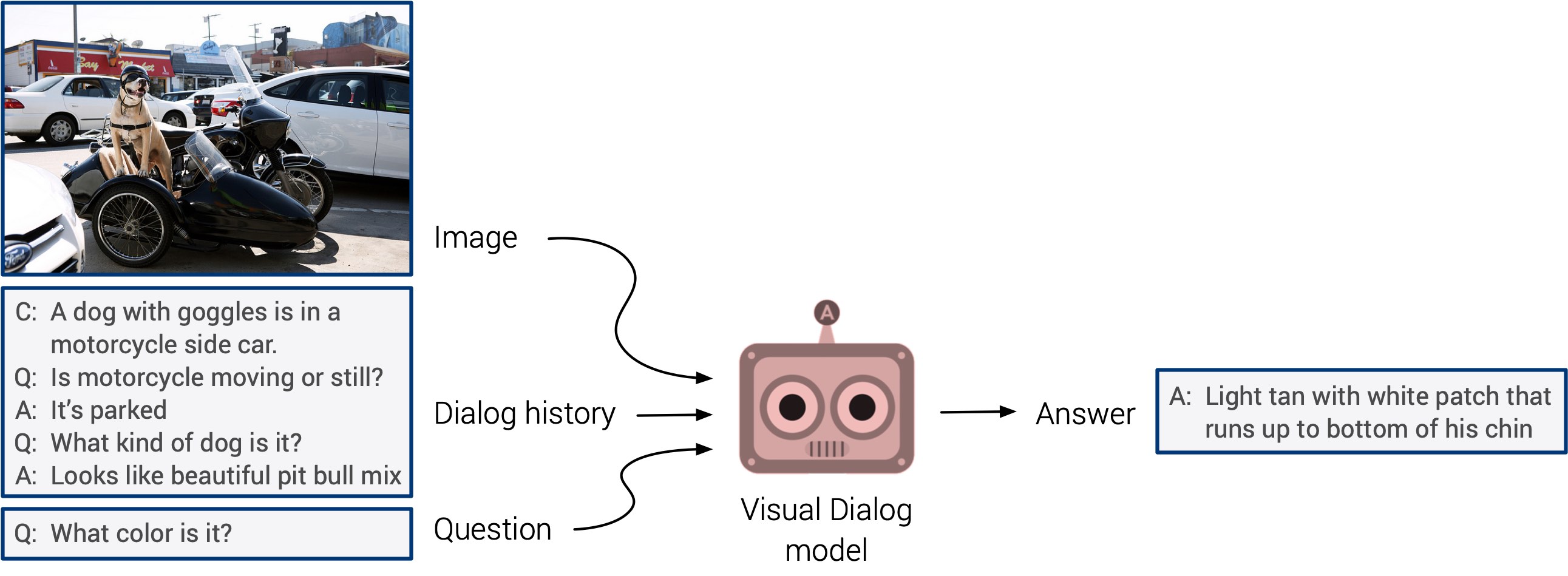}
    \caption{An example from the VisualDialog dataset. The agent has access to a reference image, to the dialogue history composed of the question/answer pairs generated so far, and it has to generate an answer for the current question. Figure adapted from \protect\cite{das2017visual}.}
    \label{fig:visdial}
\end{figure}

Agents who are able to support a conversation are much more human-like than agents that are able to answer single questions only. For this reason, the VQA task has been extended by the {\bf Visual Dialogue (VisDial)} dataset \cite{das2017visual}. In the VisDial dataset, as depicted in Figure \ref{fig:visdial}, the agent receives in input the dialogue history composed of a sequence of question/answer pairs, the current question and the current reference image. Two different tasks have been proposed for this dataset: 1) \textit{answer selection}, where the agent has to select the correct answer among a set of 100 candidates, and 2) \textit{answer generation}, where the agent should generate a response token by token. However, as highlighted by \cite{massiceti2018visual}, state-of-the-art results can be achieved by using the Canonical Correlation Analysis algorithm that relies on ad-hoc feature extractors for images (image features extracted from a pretrained \textit{ResNet} \cite{he2015resnet}) and text (question and answer representations are learned using \textit{FastText} \cite{bojanowski2017enriching}). Another downside of this task is the relevance of history. \cite{agarwal2020history} show that the number of dialogues actually requiring dialogue history is very low. Therefore, models can do well just by using the current question to generate the answer.

A dataset that proposes to give the role of first-class citizen to dialogue history and common ground is {\bf Photobook}~\cite{haber2019photobook}. It is a dataset of 2,500 human-human goal-oriented English dialogues between two participants that have to identify shared images in their \emph{photobook} by exchanging text messages. In their evaluation, they focus on the task of reference resolution and reference generation in such visually grounded dialogues. However, they do not assess the ability of artificial agents to actually play the game in a collaborative fashion by both generating utterances \emph{as well as} selecting target photos.

\subsubsection{Language Evolution and Emergent Communication}
\begin{figure}[ht]
    \centering
    \includegraphics[scale=0.4,keepaspectratio]{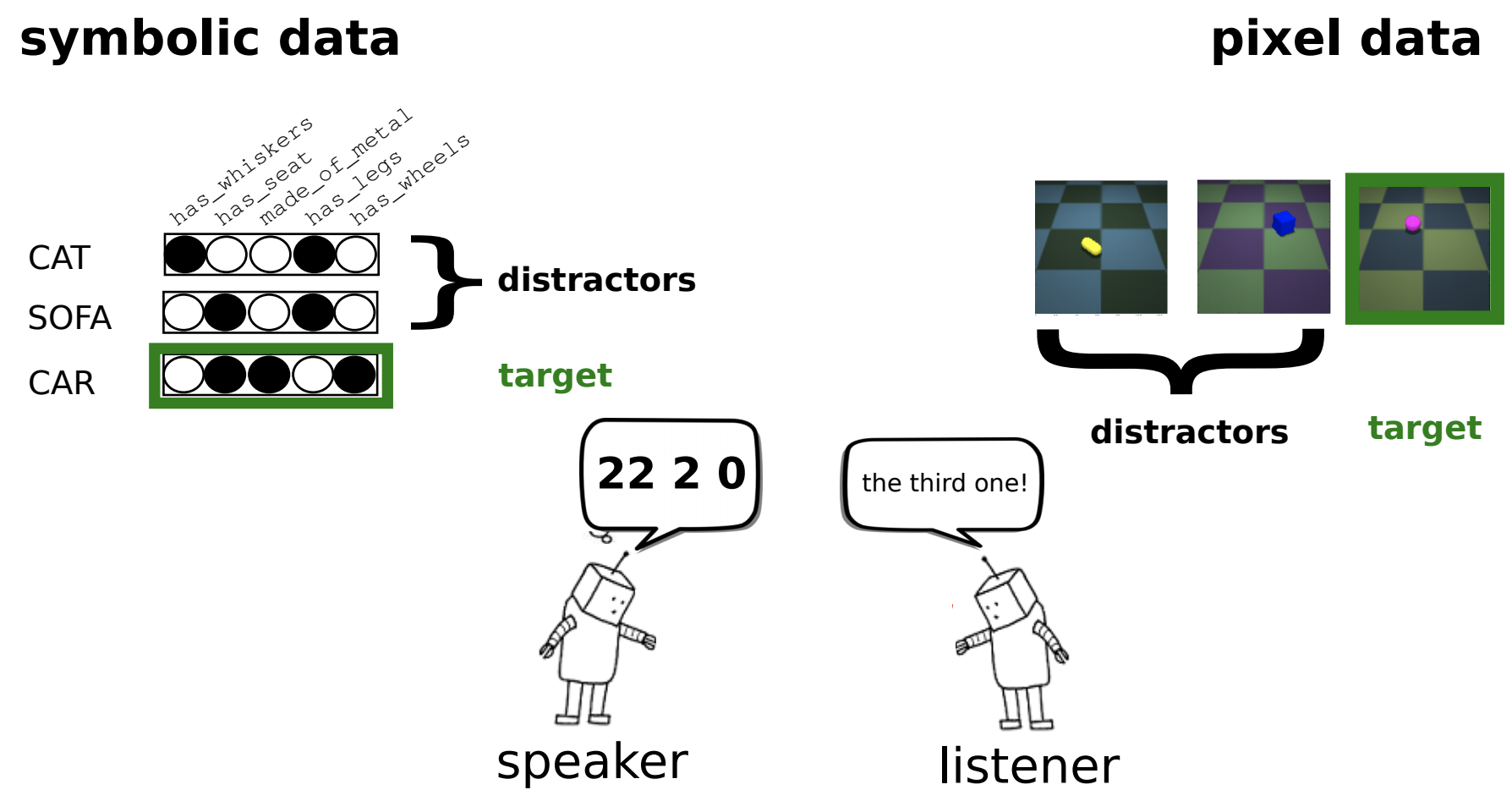}
    \caption{Referential game used by \protect\cite{lazaridou2018emergence} as a benchmark for emergent communication between two agents: the Speaker and the Listener. The Speaker, given a representation of the target image---which could be either symbolic or in the form of pixels---generates a message by sampling symbols from a fixed lexicon. Given the message sent by the Speaker, the Listener has to understand which one is the target object among the distractors. Figure adapted from \protect\cite{lazaridou2018emergence}.}
    \label{fig:ref_game}
\end{figure}

Another line of work is focused on language evolution and emergent communication in referential language games \cite{lewis1969convention}. A specific instance of the game, depicted in Figure \ref{fig:ref_game}, can be described as follows: a speaker describes an object in a scene using a set of symbols and the hearer needs to select, among a set of distractors, which one is the object described by the speaker. A recent implementation of the game was studied by several authors (e.g., \cite{lazaridou2018emergence,havrylov2017emergence,bouchacourt2018agents}) who use either synthetic images or real-world images coming from MSCOCO. A downside of these approaches is that they focus on single-turn interactions, and are thus not considering the intricacies of dialogue phenomena which are essential for an agent that has to learn by natural language interaction.
It is important to underline the fact that in the language evolution and emergent communication literature, agents start tabula rasa and form communication protocols that maximise task rewards. While this purely utilitarian framework results in agents that successfully learn to solve the task by creating a communication protocol, these emergent communication protocols do not bear core properties of Natural Language~\cite{kottur2017natural}. This is ultimately what agents are required to develop if we want them to be embodied in the environment and join in conversations with humans.

Due to the complexity introduced by having situated dialogues divided into many turns, there have been several initiatives that tried to study very specific problems in isolation. For example, \cite{clark2021iconary} proposes {\bf Iconary}, a collaborative game with two agents: 1) a Drawer that receives a phrase and has to draw it using a specific set of icons; and 2) a Guesser who generates a text to guess the phrase. This dataset is appealing because it involves an asymmetry of information that favours communication between players to guess the correct phrase. {\bf PentoRef}~\cite{zarriess2016pentoref} is another puzzle game where the Instruction Follower has to manipulate specific game pieces to achieve the goal that the Instruction Giver has in mind.
In the same spirit, \cite{kottur2019clevr} developed {\bf CLEVR-dialog}, a two-player collaborative game where the aim is to reconstruct a hidden scene graph representation known only to one of the players. The focus of this task is on the dialogue history and its importance for reference resolution. Similarly, the {\bf CODRAW} game \cite{kim2019codraw} involves two players: a \textit{Teller} and a \textit{Drawer}. The Drawer asks questions to the Teller in order to accurately reconstruct the image that only the Teller can see. The task requires that the agents develop Natural Language understanding and generation skills in order to complete the game. They use the quality of the reconstructed image as a proxy for the quality of the communicative skills of the Drawer. 
% However, similar to the VisDial task, CODRAW  assumes that the agent is already equipped with fundamental skills required to a) ``imagine" the scene and b) ask accurate questions. 

Another situated interaction dataset is  {\bf Situated and Interactive Multimodal Conversations (SIMMC)}~\cite{kottur2021simmc}. It represents a collaborative task where a user is looking for an item of clothing and has to ask an AI assistant for help in a simulated clothes shop. Another interactive situated dialogue task is {\bf CUPS}~\cite{dobnik2020local}. In this setup, the authors define a situated context represented by a table with multiple coloured cups. They assume there are multiple speakers that have different perspectives of the same scene, each with different objects missing. Their task consists of spotting which objects are missing and different. The difference in perspective favours communication between the players to succeed in the game. In this setup, we assume that the two speakers have knowledge of the language used to communicate. Instead, in {\bf BURCHAK}~\cite{yu2017burchak}, the authors collected a dataset where one of the agents, the Teacher, is trying to teach new visually grounded words in a completely made-up language to the Student, who is not familiar with it. Again, the concept of asymmetry of information is leveraged to favour communication between agents. 

\begin{figure}[ht]
    \subfigure[GuessWhat?! scene]{
    % \centering
	\includegraphics[width=0.47\linewidth]{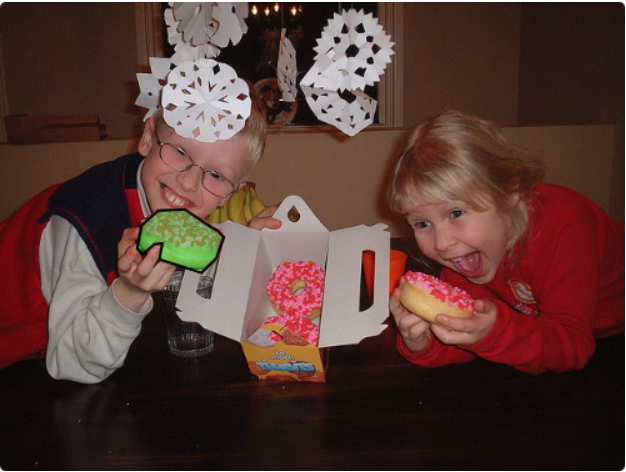}
% 	\caption{A possible reference scene in the \textit{GuessWhat?!} game. The segmented object in green (\emph{donut}) represents the target object that the \emph{guesser} needs to find out.}
    \label{fig:guesswhat_scene}
    } %
    \subfigure[GuessWhat?! dialogue]{
    %\centering
    \includegraphics[width=0.47\linewidth]{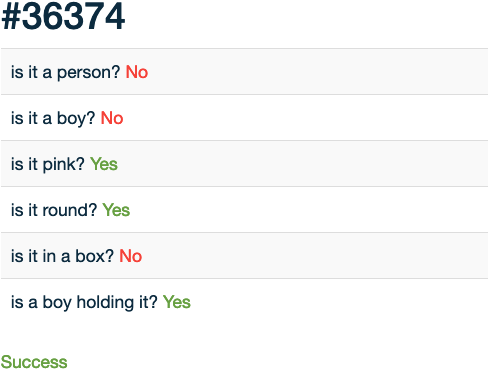}
% 	\caption{A dialogue between the \textit{questioner} and the \textit{oracle} taken from the \textit{GuessWhat?!} \cite{de2017guesswhat} dataset. The questioner needs to generate \emph{coherent} and discriminative expressions in order to collect important clues to select the target object.}
	\label{fig:guesswhat_game}
    }
    \caption{An example dialogue extracted from the \guesswhat~dataset \protect\cite{de2017guesswhat}. Figure \ref{fig:guesswhat_scene} shows a possible scene and Figure \ref{fig:guesswhat_game} shows the game played by the Questioner and the Oracle. The Oracle is aware that the target object is the ``Donut" highlighted in green, and has to support the Questioner in guessing it by replying Yes/No to their questions.}
    \label{fig:guesswhat_example}
\end{figure}

In contrast to the previously described tasks, the {\bf \guesswhat }~dataset \cite{de2017guesswhat} represents the first benchmark that involves a goal-oriented dialogue that favours: 1) the acquisition of discriminative features associated with the objects in the scene by playing the role of the Questioner -- a module which generates questions about a given image; and 2) the development of Natural Language understanding capabilities by playing the role of the Oracle -- the module which has to coherently reply to the Questioner. As shown in Figure \ref{fig:guesswhat_example}, the Questioner has to \emph{understand} the image in order to \emph{generate} coherent questions about the reference scene and, at the same time, has to \emph{understand} the feedback provided by the Oracle in order to adapt its dialogue strategy. Differently from all other tasks, \guesswhat{} involves both generation and understanding of Natural Language expressions that are grounded in the reference scene. We consider both capabilities as fundamental in learning grounded language learning. This is because agents able to support a conversation in natural language with other agents should be able to understand and generate coherent responses.

In a scenario in which agents are \textit{divergent} -- they have different perspectives, different skills, and different languages -- Natural Language communication emerges because it is essential to coordinate upon certain references, and certain goals~\cite{chandu-etal-2021-grounding-grounding,benotti-blackburn-2021-grounding}. The dataset {\bf Spot the Difference} presented by \cite{zheng2022spot} contains 95K simulated dialogues based on two agents trying to solve a referential game. Another example of divergent agents is \textbf{MeetUp!}~\cite{ilinykh2019meetup}. This is a game situated in a simulated environment where two players have to converse and coordinate in order to meet in the same room. Differently from GuessWhat?!, in this game the two agents have different visual perspectives. This asymmetry of information forces agents to communicate and clarify their references in order to complete their tasks. 

\subsubsection{Embodied Interactive Games}

\begin{figure}[ht]
    \centering
    \includegraphics[scale=0.3, keepaspectratio]{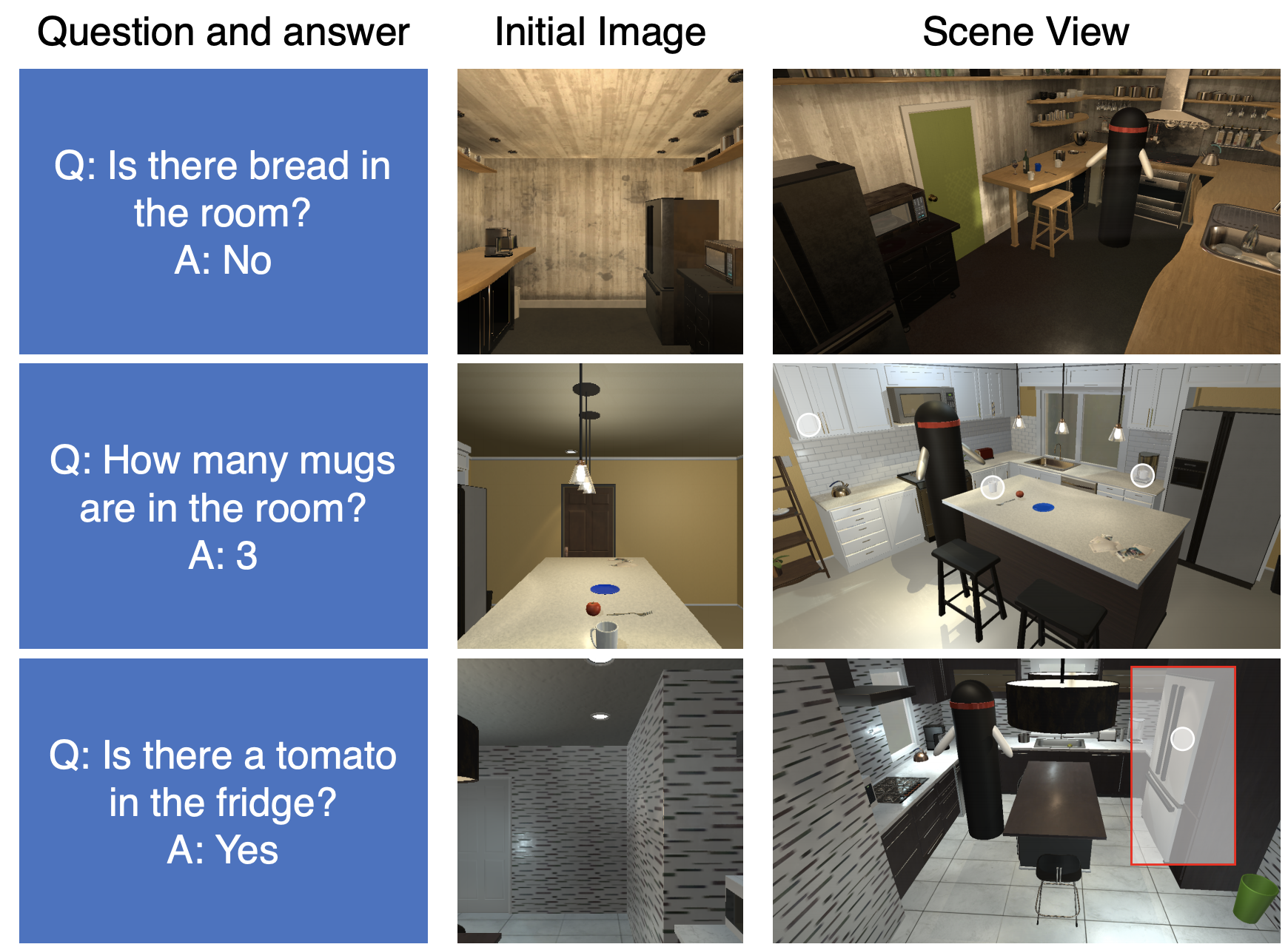}
    \caption{Example from the IQA dataset based on AI2THOR environment presented in \protect\cite{gordon2018iqa}. The agent starts from an initial position in the environment and receives the question as well. It has to move in the environment in order to find the clues that will help it to answer the question.}
    \label{fig:iqa_ai2thor}
\end{figure}

Differently from \emph{situated} tasks, many approaches can be classified as Embodied AI tasks~\cite{savva2019habitat} which are focused on the execution of actions either in the real world or within 3D virtual environments (e.g., \textit{Matterport3D}~\cite{chang2017matterport3d}, \textit{HoME} \cite{brodeur2017home}, \textit{CHALET} \cite{yan2018chalet}, \textit{HABITAT} \cite{savva2019habitat}, and
\textit{AI2THOR} \cite{ai2thor}). 
\cite{lingunet} jointly model navigation and plan execution by receiving visual information from the current scene and generating a plan of a sequence of task-dependent actions such as moving forward, changing the orientation of the camera as well as object interaction. \cite{hermann2017grounded} use the \textit{DeepMind Lab} \cite{beattie2016deepmind} environment to define a language-guided navigation task where the agent has to reach an object indicated in a natural language instruction. \cite{gordon2018iqa} present an interactive VQA task in AI2THOR (as shown in Figure \ref{fig:iqa_ai2thor}) where the agent has to answer questions related to specific objects in the environment. The agent has to navigate in the environment to discover the objects and answer questions about them. Therefore the task involves both reasoning and navigation skills to be successfully solved. A navigation task for the CHALET environment, called LANI, has been proposed by \cite{lingunet}. In addition, they propose an instruction-following dataset called CHAI that requires the agent to execute a set of actions to complete a given ``recipe" in the household domain.

Another line of work is interested in studying navigation tasks in simulated environments with photo-realistic scenes. Several datasets involving photo-realistic environments have been proposed: \cite{mirowski2019streetlearn} introduce {\bf StreetLearn} as a simulated environment to foster research in perception, planning, memory and navigation. The {\bf Room-to-Room} dataset~\cite{vln} is another relevant benchmark for navigation tasks in which the agent receives an instruction that tells it where to go and the agent has to generate a sequence of actions to reach the destination. \cite{chen2019touchdown} propose another Google Street View-based environment associated with navigation instructions that should be followed by the agent to reach a given goal position in the world. The {\bf Room-across-Room} dataset~\cite{rxr} was designed to study more fine-grained abilities of the agent to ground the instruction in temporally relevant referents of a photo-realistic 3D environment. This dataset also includes information in different languages to study how people refer to objects in space in different cultures. The dataset {\bf TalkTheWalk} \cite{de2018talk} comprises a cooperative task involving a guide and a tourist. The guide is aware of the map and has to give instructions to the tourist about how to reach a specific goal position in New York City. This dataset represents a very complex challenge for AI agents because it requires both navigation and communication skills. {\bf CerealBar}~\cite{suhr2019executing} is another collaborative game situated in a 3D environment where a follower has to collect a set of cards (with specific symbols) following the instructions of another agent. The follower has a limited set of actions that are mostly related to navigating the environment and picking up cards. Another collaborative task that requires natural language communication is {\bf MeetUp}~\cite{ilinykh2019meetup}. In this task, one agent has to follow instructions and ask for feedback to reach a room that another agent is aware of. The navigator has to provide correct spatial references so that the other agent can give instructions about where to go. A variant of this interactive navigation task has been proposed by \cite{cvdn}, namely {\bf CVDN}. It is a bigger dataset comprising 2050 human-human navigation dialogues of 7k navigation trajectories composed of question-answer exchanges across 83 houses. 

\begin{figure}[ht]
    \centering
    \includegraphics[scale=0.35]{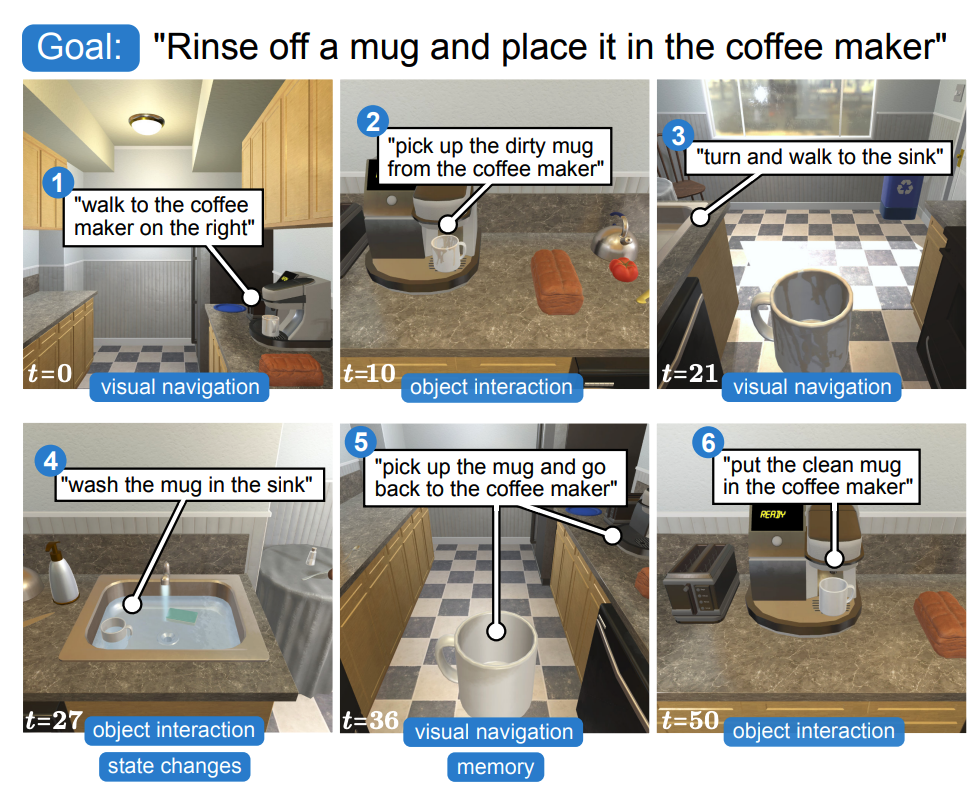}
    \caption{An example of a task to be solved in the ALFRED benchmark \protect\cite{alfred}. The agent has to master several skills to succeed such as navigation, visual memory and object manipulation. Figure adapted from \protect\cite{alfred}.}
    \label{fig:alfred}
\end{figure}

Such navigation-oriented tasks have a very important role in developing embodied agents with sophisticated reasoning and memory skills (for a more comprehensive list of these tasks refer to \cite{gu2022vision}). However, learning grounded meaning only by moving in the environment is rather limiting. In this way, the agent is not able to experience the effect of specific actions on specific objects. This is a concept that is related to the theory of affordances~\cite{gibson1950affordances} and its importance for learning grounded meanings of objects~\cite{glenberg2002grounding}. 
\cite{chevalier2018babyai} propose a virtual environment with a teacher that emits synthetic language generated by a context-free grammar. 
They define a curriculum for the agent based on increasingly complex tasks involving navigation (e.g., ``navigate to a specific object") and manipulation tasks (e.g., ``open the door on your left").
The agent has to master preliminary levels to solve successive stages of the game. The game culminates with the \textit{BossLevel} in which all the previous capabilities are required. 

Another example of a task where the agent has a richer action set is {\bf Collaborative Minecraft}~\cite{narayan2019collaborative} where an \textit{Architect} has a goal structure that wants to build and has to communicate specific instructions to the \textit{Builder} that has to move around coloured blocks to create the desired shape. In this case, the resulting dialogues are complex and rich in spatial referential expressions required to specify the position of given game pieces. An up-to-date version of this task is presented in the IGLU challenge~\cite{kiseleva2022interactive}. Another similar task that presents a similar blocks manipulation task is {\bf BlocksWorld}~\cite{bisk2016natural}. However, the set of actions that the agent is able to execute in these tasks is rather limited. The same is true for another interactive task that was proposed by \cite{abramson2020imitating}. They define {\bf Playroom}, a 3D environment where a learning agent has to perform several tasks of different complexities such as Q\&A and instruction following. Again the action set is limited to navigation actions (e.g., \emph{forward}, \emph{back}, etc.) and one manipulation action (\emph{grab}). This downside makes them an unsuitable benchmark for learning the meaning of specific actions that have an effect on the environment. \textit{AI2Thor}~\cite{ai2thor} is the only 3D environment with rich simulated physics that models a variety of actions for the agent such as \emph{pick up}, \emph{slice}, \emph{toggle}. Based on this 3D environment, the research community has proposed several challenges. For instance, {\bf ALFRED}~\cite{alfred} is the first language-guided task completion benchmark that requires both navigation and manipulation actions. As shown in Figure~\ref{fig:alfred}, the agent has to complete a goal specified by a Natural Language instruction. In addition to the overall goal instruction, instructions are provided for the agent for every sub-goal. Therefore, it can be intended as an interactive language game where the follower only executes the actions required to complete the task. In ALFRED, every task is very complex, as it is composed of very long action sequences (average trajectory length is $40$) compared to other navigation tasks (average trajectory length is around $6$). Each trajectory can be divided into a sequence of sub-goals required to achieve the final goal. Manipulating sub-goals might involve irreversible actions (e.g., slicing an apple). It is therefore essential for the agent to learn that specific actions should not be executed on specific objects. Another set of interactive tasks is presented in the {\bf SILG}~\cite{zhong2021silg} where they integrate under the same framework grid-like worlds such as NetHack~\cite{kuttler2020nethack} as well as 3D simulated environment like Touchdown~\cite{chen2019touchdown}. They all share the same symbolic representation of the scene that incredibly simplifies the visual scene that an agent is supposed to receive in input. For example, the Touchdown environment~\cite{chen2019touchdown} scenes have been converted to symbols by segmenting its panoramas into semantic grids indicating pixels that belong to specific classes (e.g., sky, bicycle, etc.). This makes the visual grounding problem even harder considering that the model is not able to learn from a variety of signals coming from real-world scenes. {\bf ALFWorld}~\cite{Shridhar2021} is another embodied AI dataset that benefits from language feedback and annotations that are automatically generated using the same game engine used for text adventure games. In this work, they stress the importance of teaching AI agents to ``imagine" the sequence of instructions that can be executed to achieve a goal, and decouple this from the actual plan execution which depends on the underlying environment. Thanks to this new environment, they demonstrate that the agent can more efficiently learn language commands in text-world environments, and reliably generalise to unseen environments.   

\begin{figure}[ht]
    \centering
    \includegraphics[scale=0.25]{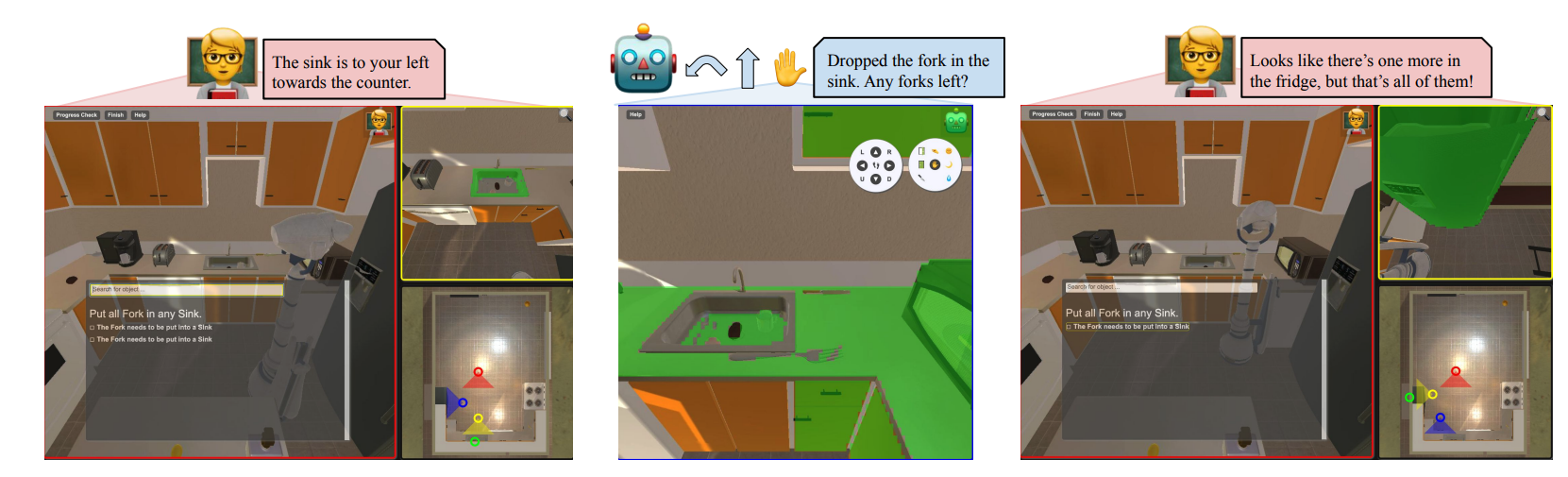}
    \caption{An example of a task to be solved in the TEACh challenge \protect\cite{padmakumar2021teach}. The Commander has to support the Follower in completing the task. The Follower is the only agent that can actually execute actions in the environment. Figure adapted from \protect\cite{padmakumar2021teach}.}
    \label{fig:teach}
\end{figure}

In the ALFRED benchmark, it is assumed that the agent has perfect vision and is always able to find all the objects required to complete the task. Unfortunately, this is not realistic and the current setup does not allow the agent to resolve ambiguities or possible misunderstandings because there is not a real interaction between the agents. In this sense, promoting a \textit{symbiotic relationship}~\cite{rosenthal2010effective} between the user and the agent, where both humans and artificial agents can solve tasks and mitigate each other's limitations is essential for language learning. Natural language offers the user a medium to flexibly express their desired outcomes and provide guidance to avoid task failure; but as we also often provide underspecified, ambiguous, or even incorrect instructions, the agent is challenged to understand known aspects of their environment and pose informative questions when uncertain, in order to minimise the risk of failure. To promote action coordination and object ambiguity resolution, \cite{gao2022dialfred} created {\bf Dialfred}, a set of 53K question/answer pairs that can be used to resolve ambiguities in the ALFRED environment. To the idea of collaboration between agents even further, \cite{padmakumar2021teach} propose the {\bf TEACh} challenge, a two-player game situated in the AI2Thor environment. Inspired by ALFRED, there are multiple tasks of different complexity to be solved. As shown in Figure~\ref{fig:teach}, there are two players with different roles: Follower and Commander. The Follower executes the instructions provided by the Commander while the Commander knows the task to be solved and provides assistance to the Follower without actually executing any actions. The ultimate challenge of this dataset is to train an agent that can support a human in completing daily tasks. Finally, to advance the state-of-the-art in Embodied Conversational AI, \cite{gao2023alexa} proposed the {\bf Alexa Arena}, a novel multi-room simulated environment that was used as the main benchmark for the Amazon Simbot challenge, a university competition to advance the state-of-the-art in Embodied Conversational AI \cite{shi2023alexa}. 

\section{Grounded Language Learning Models} \label{sec:gll_models}

In this section, we will provide a critical survey and analysis of ML models that have been proposed to tackle some of the grounded language learning tasks reviewed in the previous sections. The search of the studies was focused on the topic of grounded language learning and was limited to the time period 2015-2022. We used Scopus~\footnote{https://www.scopus.com/} to complete the search, considering only published works in relevant Machine Learning or Natural Language Processing conferences. From the original search completed in 2019 we selected a total of $32$ papers. Such papers were then extended with additional references published in major conferences in the field such as ACL, EMNLP, etc. Overall, we analysed a total of $50$ studies according to specific criteria. These criteria have been selected because we consider them essential for artificial agents able to support the visually grounded interactive language games that we described in this work. The selected criteria are reported as follows:
\begin{itemize}
    \item \textit{Compositionality}: is the representation of a complex entity the result of the combination of simpler ones learned by the model?
    \item \textit{Dialogue}: is the model able to support a conversation in natural language with another agent?
    \item \textit{Architecture}: is the proposed approach designed as an end-to-end neural architecture or as a modular architecture?
    \item \textit{Visual representation}: which kind of visual representation is used?
    \item \textit{Reasoning}: does the model support any kind of reasoning mechanism able to leverage the learned representations?
    \item \textit{Concept representation}:	which kind of representations are learned and used by the system?
    \item \textit{Few-shot learning}: Is the model able to solve a task by having access to a few learning examples?
    \item \textit{Real-world vision}: is the model exposed to real-world images?
    \item \textit{Natural Language}: is the model exposed to Natural Language (i.e., English content generated by real users)?
\end{itemize}
A summary of the comparison is reported in Table \ref{tab:gll_model_comparison}. A more detailed explanation of the work is  reported in the following section.

\begin{sidewaystable}
\resizebox{1.0\textwidth}{!}{
\begin{tabular}{|lcccccccc|}
\hline
\textbf{Paper}                     & \multicolumn{1}{l}{\textbf{Compositionality}} & \multicolumn{1}{c}{\textbf{Dialogue}} & \multicolumn{1}{c}{\textbf{Architecture type}} & \multicolumn{1}{c}{\textbf{Vision module}} & \multicolumn{1}{c}{\textbf{Concept Representation}} & \multicolumn{1}{c}{\textbf{Extreme Generalisation}} & \multicolumn{1}{c}{\textbf{Real world vision}} & \multicolumn{1}{c|}{\textbf{Natural Language}} \\ \hline
\cite{agarwal2019community}                 &                                               & yes                                   & end-to-end                                     & pretrained VGG-net features                & Distributed                                         &                                                     & yes                                            & yes                                            \\
\cite{antunes2017learn}                  &                                               & hand-crafted                          & modular                                        & multiple SOM                               & Distributed                                         & few                                                 & yes                                            & synthetic                                      \\
\cite{bahdanau2018learning}                &                                               &                                       & end-to-end                                     & CNN                                        & Distributed                                         & zero                                                &                                                & synthetic                                      \\
\cite{chou2018self}                        &                                               &                                       & end-to-end                                     & CNN                                        & Distributed                                         &                                                     & yes                                            & yes                                            \\
\cite{patki2018language}                       &                                               & hand-crafted                          & modular                                        & pipeline vision algorithms                 & Set of features                                     &                                                     & yes                                            & synthetic                                      \\
\cite{chrupala2018symbolic}                    &                                               &                                       & end-to-end                                     & pretrained VGG-net features                & Distributed                                         &                                                     & yes                                            & yes                                            \\
\cite{yu2018interactive}                      &                                               & yes                                   & end-to-end                                     & CNN                                        & Distributed                                         & zero                                                &                                                & synthetic                                      \\
\cite{shekhar2019beyond}                   &                                               & yes                                   & end-to-end                                     & pretrained VGG/ResNet152                   & Distributed                                         & zero                                                & yes                                            & yes                                            \\
% de Vries (2018)                    &                                               & yes                                   & end-to-end                                     &                                            & Distributed                                         &                                                     &                                                & yes                                            \\
\cite{pillai2018unsupervised}                      &                                               &                                       & modular                                        & RGB colour                                 & visual classifiers                                  &                                                     & yes                                            & synthetic                                      \\
\cite{deng2018visual}                        &                                               & yes                                   & end-to-end                                     & pretrained VGG                             & distributed                                         &                                                     & yes                                            & yes                                            \\
\cite{yi2018neural}                          & yes                                           &                                       & modular                                        & Mask-CNN                                   & Distributed, attribute-based                        & zero                                                &                                                & yes (only one dataset)                         \\
\cite{das2018embodied}                         &                                               &                                       & modular                                        & CNN                                        & distributed                                         &                                                     & 3D env                                         & synthetic                                      \\
\cite{williams2018learning}                    & yes                                           &                                       & modular                                        &                                            & lambda calculus                                     &                                                     & yes                                            & yes                                            \\
\cite{chevalier2018babyai}  &                                               &                                       & end-to-end                                     & CNN + FiLM                                 & distributed                                         &                                                     &                                                & synthetic                                      \\
\cite{co2018guiding}                 &                                               &                                       & end-to-end                                     &                                            & distributed                                         &                                                     &                                                & synthetic                                      \\
\cite{lu2018neural}                       &                                               &                                       & end-to-end                                     & CNN, Mask CNN                              & Distributed                                         &                                                     & yes                                            & yes                                            \\
\cite{zhang2018oneshot}                       &                                               & yes                                   & end-to-end                                     & CNN                                        & Distributed                                         & few                                                 &                                                & synthetic                                      \\
\cite{mordatch2018emergence}                    & yes                                           & multi-agent communication             & end-to-end                                     &                                            &                                                     &                                                     &                                                &                                                \\
\cite{shah2018follownet}                        &                                               &                                       & end-to-end                                     & CNN (depthmap and segmentation map)        & Distributed                                         &                                                     & yes                                            & synthetic                                      \\
\cite{misra2018learning}                    &                                               & yes                                   & modular                                        & CNN resnet                                 & Distributed                                         & zero                                                &                                                & yes (only one dataset)                         \\
\cite{kiela2018learning}                    &                                               &                                       & end-to-end                                     & CNN                                        & Distributed                                         &                                                     & yes                                            & yes                                            \\
\cite{kiros2018illustrative}                    &                                               &                                       & end-to-end                                     & CNN                                        & Distributed                                         &                                                     & yes                                            & yes                                            \\
\cite{becerra2018gold}                  & yes                                           &                                       & Prolog-based                                   & prolog-representation hand-crafted         & Horn clauses                                        &                                                     &                                                &                                                \\
\cite{kalyan2018learn}                      &                                               &                                       & end-to-end                                     & CNN                                        & Distributed                                         &                                                     & yes                                            & yes                                            \\
\cite{yu2018guided}                      &                                               &                                       & end-to-end                                     & CNN                                        & Distributed                                         &                                                     &                                                &                                                \\
\cite{thomason:icra19}                & yes                                           & yes                                   & modular                                        & visual classifiers                         & visual classifiers                                  &                                                     & yes                                            & yes                                            \\
\cite{choi2018compositional}                     & yes                                           &                                       & end-to-end                                     & CNN                                        & distributed                                         & zero                                                &                                                & emerged language                               \\
\cite{xie2018visual}                    &                                               &                                       & end-to-end                                     & CNN or Mask-CNN                            & distributed                                         &                                                     & yes                                            & yes                                            \\
\cite{gordon2018iqa}                  &                                               &                                       & end-to-end                                     & CNN                                        & distributed                                         &                                                     & 3D env                                         & synthetic                                      \\
\cite{kottur2019clevr}                      & yes                                           & yes                                   & end-to-end                                     & CNN                                        & distributed                                         &                                                     & synthetic                                      & synthetic                                      \\
\cite{zhou2020vlp}                     &                                               &                                       & end-to-end                                     & FastRCNN                                   & distributed                                         & both                                                & yes                                            & yes                                            \\
\cite{majumdar2020improving}     &                                               &                                       & end-to-end                                     & CNN                                        & distributed                                         &                                                     & yes                                            & yes                                            \\
\cite{zhang2021vinvl}     &                                               &                                       & end-to-end                                     & CNN                                        & distributed                                         &                                                     & yes                                            &                                                \\
% Xu, Haiyang, et al. (2021)         &                                               &                                       & end-to-end                                     & CNN                                        & distributed                                         &                                                     & yes                                            & yes                                            \\
\cite{et}                &                                               &                                       & end-to-end                                     & CNN                                        & distributed                                         &                                                     & 3D env                                         & yes                                            \\
\cite{min2021film}         &                                               &                                       & modular                                        & hybrid                                     & hybrid                                              &                                                     & 3D env                                         & yes                                            \\
\cite{crossmap}                &                                               &                                       & end-to-end                                     & CNN                                        & distributed                                         &                                                     & 3D env/photorealistic                          & yes                                            \\
\cite{lxmert} &                                               &                                       & end-to-end                                     & FastRCNN                                   & distributed                                         & both                                                & yes                                            & yes                                            \\
\cite{zhang2021visually}           & yes                                           &                                       & modular                                        & FastRCNN                                   & distributed                                         & both                                                & yes                                            & yes                                            \\
\cite{kuo2020compositional}        & yes                                           &                                       & modular                                        & CNN                                        & distributed                                         &                                                     & synthetic                                      & no                                             \\
\cite{jiang2019language}       & yes                                           & no                                    & end-to-end                                     & CNN                                        & distributed                                         & both                                                & synthetic                                      & no                                             \\
\cite{zellers2021piglet}    &                                               &                                       & modular                                        & attribute-representation                   & distributed                                         & few                                                 & 3D env                                         & yes                                            \\
\cite{zellers2021merlot}    &                                               &                                       & end-to-end                                     & CNN                                        & distributed                                         &                                                     & yes                                            & yes                                            \\
\cite{singh2021illiterate}           & yes                                           & no                                    & end-to-end                                     & CNN                                        & distributed                                         & few                                                 & synthetic                                      & no                                             \\
\cite{radford2021learning}       &                                               & no                                    & end-to-end                                     & CNN                                        & distributed                                         & both                                                & yes                                            & yes                                            \\
\cite{ramesh2021zero}      &                                               & no                                    & end-to-end                                     & CNN                                        & distributed                                         & both                                                & yes                                            & yes                                            \\
\cite{lazaridou2018emergence} &                                               & no                                    & end-to-end                                     & CNN                                        & distributed                                         & few                                                 & synthetic                                      & emerged language                               \\
\cite{lazaridou2016multimodal} &                                               & no                                    & end-to-end                                     & CNN                                        & distributed                                         &                                                     & yes                                            & yes                                            \\
\cite{lazaridou2016multiagent}            &                                               & no                                    & end-to-end                                     & CNN                                        & distributed                                         &                                                     & yes                                            & emerged language                               \\
\cite{lazaridou2020multi}            &                                               & no                                    & end-to-end                                     & CNN                                        & distributed                                         &                                                     & synthetic                                      & yes                                            \\
\cite{zhu2021few}                  & no                                            & no                                    & end-to-end                                     & CNN                                        & distributed                                         &                                                     & 3D env                                         & no                                             \\ \hline
\end{tabular}
}
\caption{Comparison of several multimodal models developed for several grounded language learning tasks.}
\label{tab:gll_model_comparison}
\end{sidewaystable}

From the completed literature review it emerged that in studying grounded language learning, often one of the two modalities (i.e., vision and language) is lacking or is not considered at all. Indeed, several of the analysed papers either use real-world vision or synthetic images while focusing on more complex language structures. Bridging this gap is really important in order to understand how well current neural models are able to handle complex visual scenes, as well as interesting Natural Language phenomena available in human language annotations. Running experiments in the real world would be ideal to simulate such complex visual scenes---although this is not always practical. One way to cope with this limitation is by designing rich 3D environments that expose models to photo-realistic visual scenes. Instead, when analysing the language dimension, another simplification is that the tasks typically involve discriminative or generative tasks involving a single interaction. Only very few approaches consider interactive tasks where the dialogue history is actually relevant for the task at hand \cite{agarwal2020history}. In some cases, the developed models are actually able to learn a dialogue strategy to complete the task whereas others assume that the dialogue is somehow scripted in advance via a hand-crafted procedure. 

\subsection{Learned Multi-modal Representations}

We are interested in understanding the type of learned representations these systems are learning, and the way they are using them for specific grounded language learning tasks. 

\subsubsection{Visual representations}

A few approaches rely on low-level sensory features modelling RGB colour channels or shape information\cite{antunes2017learn,patki2018language}. Based on the assumption that an image can be expressed as a bag of visual words~\cite{csurka2004visual,grauman2005pyramid,sivic2005discovering}, several works (e.g., \cite{thomason:icra19,yu2017learning}) use a set of visual classifiers that are manually mapped to specific classes that represent visual concepts. Due to the expressivity of deep neural networks, a common practice is to use latent features learned by a Convolutional Neural Network (CNN) trained on a large-scale image dataset (e.g., ImageNet~\cite{deng2009imagenet}). Specifically, the features belonging to the last layer of a CNN, have been widely adopted due to their ability to be re-used for other tasks other than image classification. With such features, it is possible to represent an entire scene with a single distributed representation. However, as demonstrated in the VisDial challenge, using an object detector to predict object bounding boxes (e.g., \textit{FastRCNN}~\cite{anderson2018bottom}) really improves performance. Having such object-centric representations allows the agent to learn the task by reasoning over objects and their relationships which are potentially invariant across scenes. In this setup, the model receives in input a high-dimensional feature vector for every detected object (Region Of Interest (ROI) vector). Additionally, to provide the model with positional information about the objects, bounding box coordinates are encoded as additional features. To incorporate class information associated with each object, either class labels are encoded~\cite{lxmert} or a probability distribution over the object detector label set~\cite{zhou2020vlp}. This provides the model with more fine-grained information associated with the scene as well as a better representation of the objects in it. This approach was used by several recent models (e.g., \cite{xie2018visual,lu2018neural,yi2018neural,lxmert,zhou2020vlp}). However, extracting features from an object detector is a very time-consuming operation and limits the ability of the model to encode specific objects that are not part of the object detector training set. Therefore, some approaches revisited grid features generated by a CNN to improve the efficiency of multi-modal models (e.g., \cite{jiang2020defense,kim2021vilt}). 

\subsubsection{Language representations}
Language representations, on the other hand, are typically trained separately, either using the reference task dataset or by pretraining using self-supervised techniques. Such representations are either in the form of word embeddings (e.g., \cite{mikolov2013linguistic,pennington2014glove}) or contextual embeddings (e.g., \cite{devlin2018bert}). The latter have had tremendous success in both Natural Language Understanding, and Natural Language Generation tasks (e.g., GPT-3~\cite{brown2020language}).

\subsubsection{Vision+Language representations}
Once we have a representation for each modality, how do we derive a \textit{multimodal representation}? A multimodal representation is a representation of data that is obtained by combining multiple sources of input such as visual, perceptual or symbolic information~\cite{baltruvsaitis2018multimodal}. Representing multiple modalities poses many difficulties regarding the way we combine heterogeneous sources and how to deal with noise in them (i.e. when they are just noisy or completely missing). Therefore, designing mechanisms to meaningfully represent such data becomes a crucial problem in multimodal representation learning.
In recent work by \cite{baltruvsaitis2018multimodal}, several multimodal fusion mechanisms have been reviewed:
\begin{itemize}
    \item \textbf{early}: the model learns multi-modal representations as part of the learning objective by jointly minimising a loss function that optimises a posterior involving both modalities;
    \item \textbf{middle}: the model uses independent training data and objectives for specific modalities. A parameterised similarity function can be used to  learn the relationships between features in multi-modal spaces;
    \item \textbf{late}: it is the converse of the previous approach because it first computes the similarity scores between the representations and then it aggregates them together.
\end{itemize}
As highlighted by \cite{kiela2017phd}, a combination between \emph{early} and \emph{middle} fusion can be considered plausible from a cognitive perspective because ``it allows for learning unimodal representations independently but which also
allows for combining said representations into an overall multimodal one that takes all modalities into account, possibly in varying degrees".

For instance, early fusion models have been relatively successful in multimodal tasks thanks to the development of large-scale unimodal encoders. Concatenating, averaging or summing unimodal representations are some of the methods used to generate multimodal representations (e.g., \cite{shekhar2019beyond,das2018embodied}). More powerful multimodal architectures have been proposed that use specific co-attention mechanisms (e.g., \cite{deng2018visual,yu2018guided,lu2018neural}). More recently, Transformers have been used to learn multimodal representations by combining modality-specific inputs via self- and cross-attention. Such models are indeed closer to the early/middle fusion models identified by \cite{kiela2017phd}.
%(e.g., \cite{zhou2020vlp, lxmert, clip, ramesh2021zero}). 

Transformer-based models have been particularly successful in most of the grounded language learning tasks that we reviewed. Following \cite{bugliarello2021multimodal}, we can define two different types of multimodal architectures: 1) \textit{single-stream} Transformers; and 2) \textit{dual-stream} Transformers. Single-stream Transformers assume that there is a single Transformer stack that receives in input both visual and textual inputs. Dual-stream Transformers instead assume that each modality is encoded by a modality-specific encoder whose outputs are fused by using a dedicated cross-modal encoder. This distinction was similarly described by \cite{baltruvsaitis2018multimodal} as \textit{joint encoder representations} and \textit{coordinated encoder representations}, respectively. However, \cite{bugliarello2021multimodal} demonstrates dual-stream attention functions act as restricted versions of the attention function in any single stream Transformer architecture. So dual-stream architectures have an inductive bias that favours interactions between modalities.

\begin{figure}
    \centering
    \includegraphics[scale=0.25]{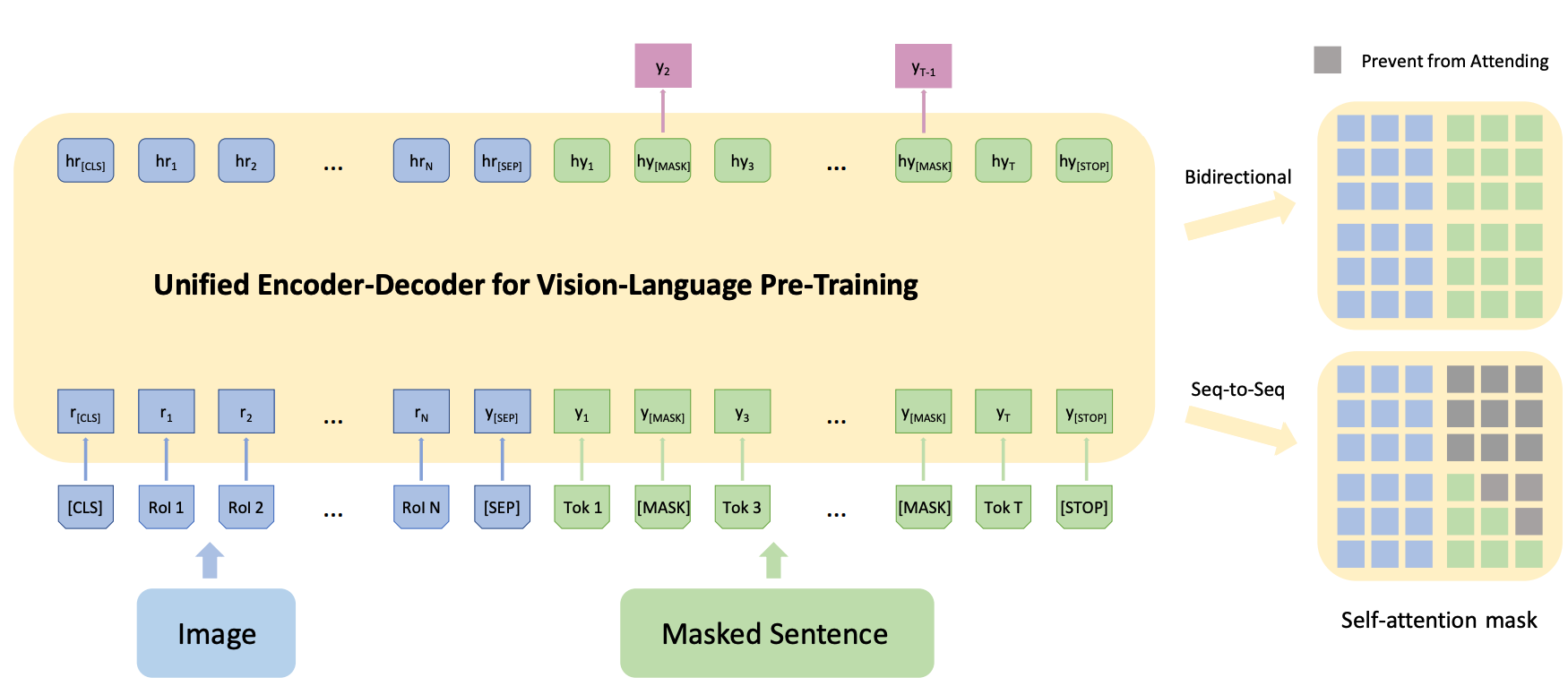}
    \caption{An overview of the VLP architecture presented by \protect\cite{zhou2020vlp}. It uses a single Transformer stack to encode both visual tokens and textual tokens. Then, it is pretrained on Conceptual Captions using both masked language modelling and sequence-to-sequence loss. Figure adapted from \protect\cite{zhou2020vlp}.}
    \label{fig:vlp}
\end{figure}

\paragraph{Single-stream Multimodal Transformers}
One of the first single-stream multimodal models, together with many others (e.g., \cite{uniter,visualbert,oscar}), is VLP~\cite{zhou2020vlp}. Its architecture is depicted in Figure~\ref{fig:vlp}. First of all, the architecture mimics the one used by BERT~\cite{bert}. Following BERT, it represents the language input as Wordpieces~\cite{schuster2012japanese}. It introduces \textit{visual tokens} which are derived from object-centric representations generated by FastRCNN. Such visual tokens are then projected to a dimensionality comparable to textual tokens. In this way, they can be passed in input to the stack of Transformer layers to generate multimodal representations of the input. Following the success of self-supervised learning for text-based Transformers, VLP is initialised with a pretrained BERT and then is pretrained using several multimodal losses on the Conceptual Captions dataset~\cite{sharma2018conceptual}. In order to support both discriminative and generative tasks, VLP is trained with both masked language modelling loss as well as with a sequence-to-sequence loss (they are randomly alternated during training). In this way, it is one of the few models that can be used for both VQA and image captioning. Other models have also proposed additional self-supervised losses for the visual modality as well. For instance, \textit{UNITER}~\cite{uniter} uses masked region modelling which consists in reconstructing the original visual tokens or predicting their class. By learning to predict missing objects, the model learns to model the visual context in a better way which can enhance language understanding.  Another model that belongs to this family is FLAVA~\cite{singh2022flava}, which takes advantage of the advancements in Vision Transformers~\cite{dosovitskiy2020image} to learn representations of the image directly from pixels without any object detector features. Both language and visual features are then fed into modality-specific Transformers that derive representations for each modality. Then, a single cross-modal Transformer derives multimodal representations. The main novelty of FLAVA is represented by its ability to perform both multimodal and unimodal tasks. In particular, it is pretrained with both multimodal and unimodal datasets so that it can simulate scenarios where only some modalities are available. Additionally, it can leverage the additional signal that is derived from both unimodal and multimodal task-specific losses.

\paragraph{Dual-stream Multimodal Transformers}

\begin{figure}
    \centering
    \includegraphics[scale=0.4]{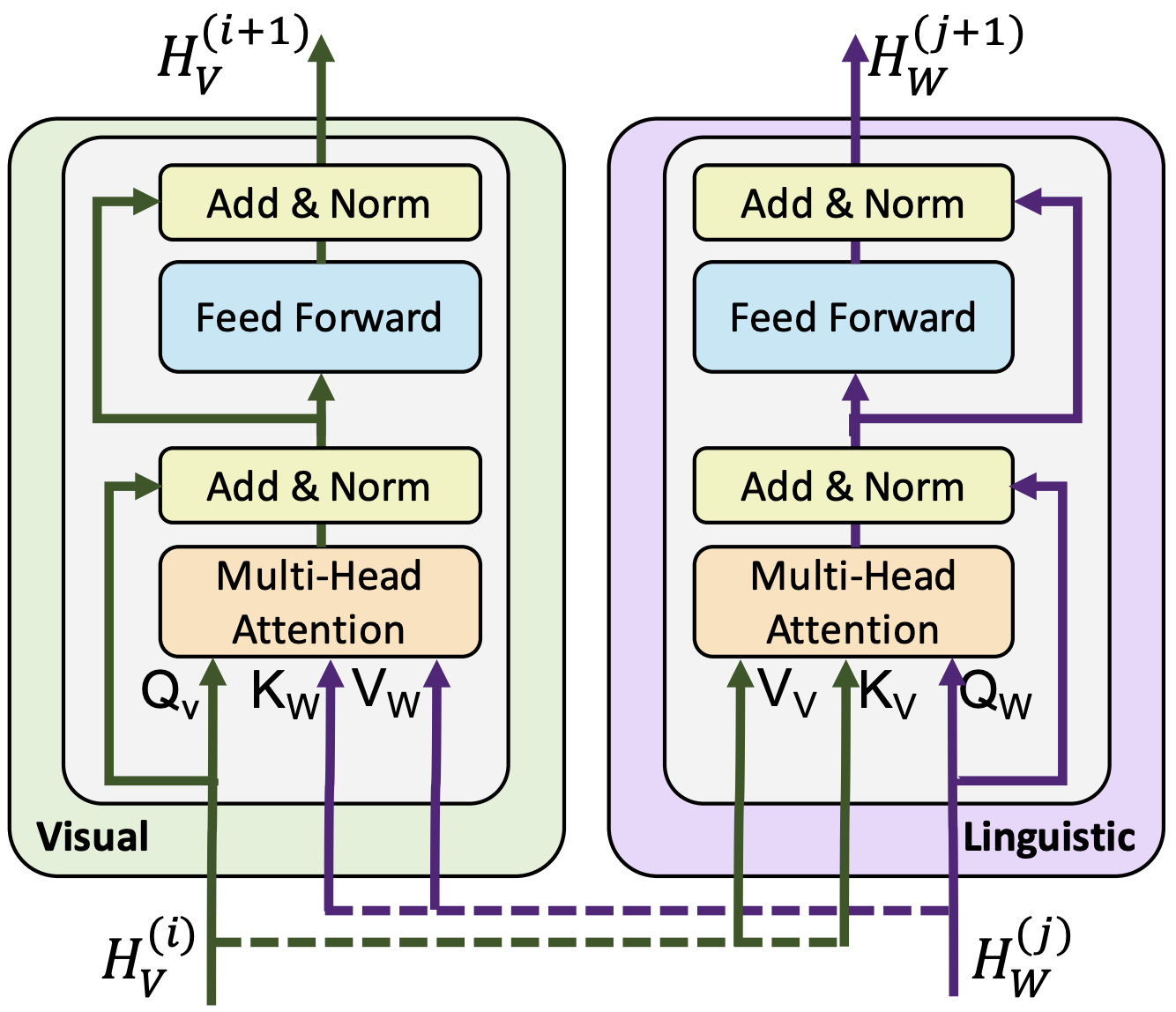}
    \caption{The co-attention layer implemented in ViLBERT to fuse vision and language representations. The co-attention layer relies on multi-head self-attention, and is applied in parallel to both vision and language representations to derive multimodal representations. Figure adapted from \protect\cite{vilbert}.}
    \label{fig:vilbert_co_attention}
\end{figure}

In contrast with single-stream multimodal Transformers, many dual-stream transformers have been proposed in the literature. As highlighted before, these models have two dedicated modality-specific encoders that are used to encode language and vision separately. An additional cross-modal Transformer block is then used to ``fuse" the hidden representations generated for the two modalities. One instance of these models is LXMERT~\cite{tan2019lxmert} which has a ``Object-relationship Encoder", a dedicated Transformer block that encodes object representations derived from a pretrained object detector, and a ``Language encoder" implemented as a Transformer block that encodes language tokens. Additionally, a ``Cross-modality Encoder", implemented as another Transformer block, is used to fuse the two modalities and favours the development of cross-modal representations that are used for the downstream tasks. This process is performed in a symmetric way. First, language representations are used as queries in the self-attention operations against the vision token representations. Then, vision token representations are used as queries in the multi-head self-attention operation. Concurrently, ViLBERT~\cite{vilbert} defined an alternative way of using cross-attention to ``fuse" the vision and language modalities. This is implemented as a two-step approach where, after using modality-specific encoders to generate latent representations for language and vision respectively, the model uses a `co-attention` transformer layer (showed in Figure~\ref{fig:vilbert_co_attention}) for a target modality that performs multi-head self-attention using as queries the representations from the target modality, and as keys and values the representations associated with the other modality.  

\subsection{Compositionality}

Natural Languages are characterised by immense \textit{productivity} meaning that they can generate an infinite number of expressions. According to linguists, this is possible thanks to the property of compositionality. Of specific interest is \textit{semantic compositionality}, the principle whereby the meaning of a linguistic expression is a function of the meaning of its components and the rules used to combine them \cite{frege1892sinn,montague1970universal}. In this paper, we intend a broader definition of compositionality that is related to the ability to combine simpler features to obtain novel objects/entities. This broader definition can be extended to compositionality in vision where an object can be seen in terms of its attributes (i.e., cat = has fur, has eyes, has 4 legs, etc.). When studying the generalisation power of artificial agents, it is fundamental to consider compositionality because it can be conjectured to be a landmark not only of language but of human thought in general \cite{fodor1988connectionism,lake2017building}. Following \cite{harnad1990symbol}, this broader definition of compositionality is the key ingredient of learning grounded representations. As demonstrated in the GROLLA evaluation framework~\cite{suglia-etal-2020-compguesswhat}, learning grounded representations should result in high performance for in-domain examples as well as generalisation to examples never seen during training.

Several papers take into account only some aspects of compositionality which are not fully integrated into the learned representation nor in the model reasoning mechanism. The work presented by \cite{choi2018compositional} studies how, in a referential game, two agents evolve their linguistic skills to generate a compositional language. In particular, they develop a two-player image description game: the speaker and hearer receive two random images. The speaker needs to generate a sequence of characters to describe the image that they see. The hearer looks at its image and needs to decide if they are looking at the same image or not. They base their model on the \textit{Obverter model} \cite{batali1998computational}. The obverter technique motivates an agent to search over messages and generate the ones that maximise their own understanding. They study the properties of the language that evolved from the communication game. Additionally, they complete a zero-shot evaluation by holding out five objects from the dataset during the training and observing how agents describe them during the test phase. Discriminative features in this case are just colour and shape (with a strong bias towards the colour) so the task cannot tell to what extent these networks are able to compose the learned representations. \cite{mordatch2018emergence} study the properties of the evolved communication protocol in a population of agents that can interact with each other via either verbal or non-verbal communication (i.e., pointing or guiding). The environment contains a fixed number of agents and landmarks. Every agent can either execute an action (i.e., go to location, look at location and do nothing) or utter a symbol. Every agent has a goal that is not revealed to the other agents requiring communication skills in order to be successful. 

\begin{figure}
    \centering
    \includegraphics[scale=0.35]{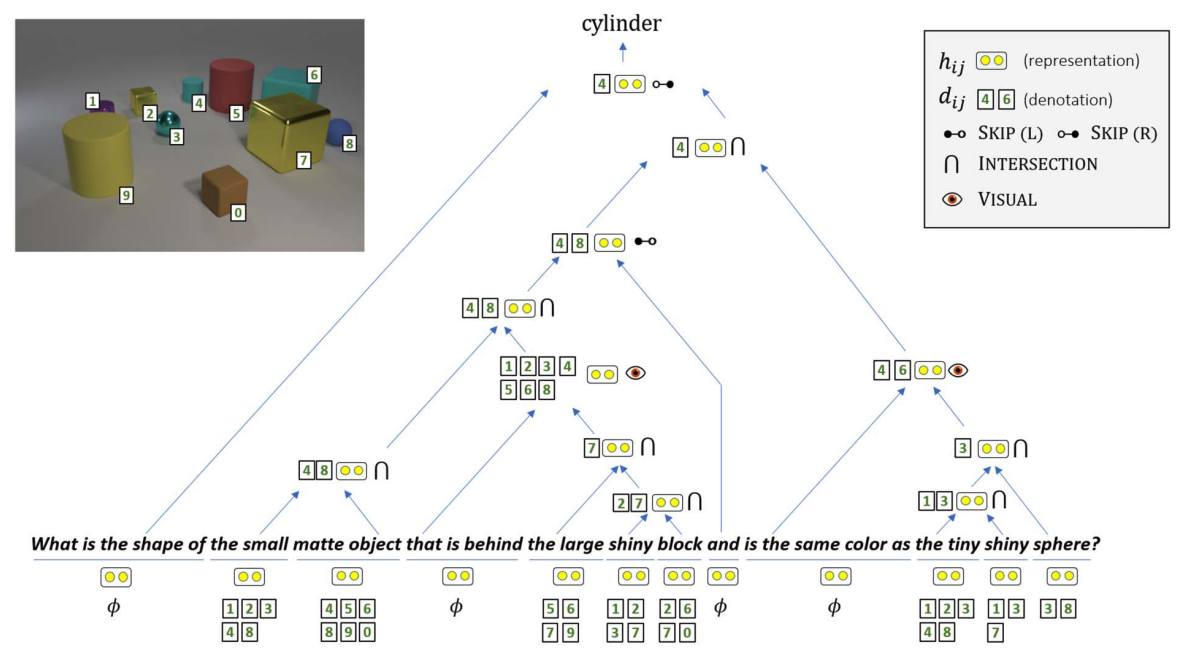}
    \caption{The Neural compositional module developed by \protect\cite{bogin2021latent}. It uses a CKY-like algorithm to derive, in a bottom-up fashion, a compositional representation of the question that is then used, via its denotation, to generate the answer. Figure adapted from \protect\cite{bogin2021latent}.}
    \label{fig:module_net_vqa}
\end{figure}

Differently from these models that try to analyse to what extent neural networks are able to learn grounded compositional language, the work presented by \cite{thomason:icra19,kottur2019clevr,yi2018neural} includes logical form as a structural bias in their models. \cite{thomason:icra19} uses a semantic parser---separately trained from the agent---to derive semantic representations of the language in the form of predicative structures. 

Powerful generalisation capabilities can be obtained when compositional representations are directly integrated in the model. For instance, \cite{yi2018neural} presents a model that learns to answer questions about images by ``executing" a sequence of predefined functional programs. First, it learns to generate a rich scene graph for the reference image and, as a second step, generates a functional program for the given question that, once executed against the scene graph, returns the answer. \cite{kottur2019clevr} uses CorefNMN \cite{kottur2018visual} for the CLEVR-dialog task with an approach similar to \cite{yi2018neural}. CorefNMN is a neural model equipped with a \textit{program decoder} that generates functional programs given the dialogue history and the current question. Then, it derives the answer by executing the extracted functional programs. A specific program ``Refer" can be used by the model to have access to a pool of previously mentioned entities in order to complete visual coreference resolution over the turns of the dialogue. Integrating these functional programs in the model guarantees a high level of generalisation because the model does not focus on learning spurious correlations between the dataset but it learns to effectively use the provided ``modules" to complete the task. 

Relying on ad-hoc symbolic representations to facilitate the generation of compositional representations was proposed first by \cite{socher2011parsing} and then revisited by \cite{bogin2021latent}. As shown in Figure~\ref{fig:module_net_vqa}, \cite{bogin2021latent} use a CKY-based algorithm~\cite{kasami1966efficient,younger1967recognition} that starts by grounding the single word embeddings associated to the words in the question. After this step, in a bottom-up fashion, compositional representations are induced from sub-span latent representations. They use a weighted sum of sub-span representations to make the computation differentiable. Each sub-span weight is derived as an attention score that represents how suitable that span is for a given context. Similar in spirit, a denotation for each span is used to derive the meaning of the sub-expression, and ultimately to derive the answer. Following the work by \cite{andreas2016neural}, they have several hand-crafted modules for executing operations on the object set that the model learns to execute. This model seems to do very well on the synthetic dataset CLEVR~\cite{johnson2017clevr} but, when tested on questions written by humans, it generalises marginally better than a fully neural counterpart (e.g., \textit{MAC}~\cite{hudson2018compositional}). This might be because some operators (e.g., computing the maximum value in a set) cannot be obtained by composing elements in the question, but instead, require higher-order reasoning. The other problem is that more sophisticated grounding is required in some cases. Their model only grounds specific phrases to objects. However, in some cases, more fine-grained properties or numbers have to be grounded which is something that the model doesn't support. 
%Therefore, combining such an approach with more flexible multimodal architectures that are not constrained by the implemented sub-modules might be an interesting avenue for future work. 
The same \textit{ground-compose} approach was implemented in the work presented by \cite{zhang2021visually}. They apply a syntactic parser on a caption to generate a symbolic representation that will guide the composition of grounded elements. First, specific noun phrases are grounded via dedicated modules to obtain a visually-grounded representation. Then, the symbolic representation is used to combine such grounded representations together. 

All of the analysed papers rely on distributed representations for the words that completely ignore the extralinguistic information associated to a concept and only capture information about the use of the word in context \cite{westera2019don}. In fact, these models do not have a representation for the concept of \texttt{cat} that defines a cat as an animal with 4 legs, a nose, and coloured fur, etc. Even multi-modal models can partially incorporate in their latent representation information related to the colour and to specific shapes that can be associated with specific concepts. However, these models do not seem to be able to take into account more complex attributes that people learn about concepts. In this sense, a compositional representation of \texttt{cat} should be derived by integrating specific sensory-motor properties (e.g., visual, auditory, etc.) as well as situational properties (e.g., affordances) that uniquely characterise a cat. Additionally, because such learned representations are typically encoded in a single vector representation, it is harder to understand what the relevant factors of variations are, and how they have been combined together. For instance, \cite{gardenfors2014geometry} in its \textit{Conceptual Spaces} theory defines an \textit{object category} as composed of domains, a set of convex regions in such domains, and information about mereonomic (part-whole) relations as well as relations between dimensions across different domains. It is very possible that distributed representations derived from CNNs are able, for example, to extract information about object colours, which resemble one possible domain identified in \cite{gardenfors2014geometry}'s theory. However, it is not clear to what extent distributed representations are expressed in terms of relational information. Such relational knowledge can be intended as the essence of compositionality because it would allow representations to be derived by composing simpler ones together.

\subsubsection{Assessing compositionality}

Despite neural networks' impressive performance on several benchmarks, many researchers have advocated that their representations and reasoning mechanisms are still very brittle \cite{marcus2018deep}. As stated in \cite{linzen2020can}, current Deep Learning models are evaluated using a protocol that does not assess their ability to generalise to out-of-distribution data. This is a capability akin to systematic generalisation, and is therefore a core property of compositionality. Previous work on visual question answering (VQA) measures generalisation to novel question-answer pairs generated for natural~\cite{agrawal2017cvqa,whitehead2021separating}, or synthetic images \cite{bahdanau2018systematic,johnson2017clevr}. Similarly, \cite{suglia-etal-2020-compguesswhat}, proposes an evaluation framework that augments a goal-oriented evaluation metric with two auxiliary tasks aimed at assessing systematic generalisation. Prior work has examined systematic generalisation for image captioning \cite{atzmon2016learning,nikolaus2019compositional,bugliarello2021role}. \cite{ruis2020benchmark} examines compositionality in the context of language-guided action execution in grid world environments extending the framework developed by \cite{lake2019compositional}.

Systematic generalisation is not the only property that a compositional system can demonstrate. For instance, \cite{pantazopoulos2022combine} extends the multi-faceted compositional evaluation framework presented by \cite{hupkes2020compositionality} to image captioning. According to their definition, an image captioning model can be considered \textit{compositional} if possesses 3 different properties: (1) \textit{systematicity}: the ability to generalise to unseen combinations of concepts learned in isolation during training; (2) \textit{productivity}: the capacity to extend predictions beyond the training observations; and (3) \textit{substitutivity}: the robustness of predictions under synonym substitution.

By analysing a model under the above-mentioned compositional frameworks, we can assess its ability to be robust to unseen examples. Therefore, concepts such as compositionality and robustness, are connected to the concept of visual grounding. In the context of visual question answering, there have been several other works that have tested the robustness of V+L models. For instance, \cite{hudson2019gqa} proposes a VQA benchmark that augments the VQA accuracy with robustness measures such as \textit{plausibility}, \textit{validity}, and \textit{consistency}. The work presented in \cite{thrush2022winoground} assesses the ability of the model to discriminate correct (image, caption) pairs from foil ones that are obtained by rearranging nouns and adjectives. CrossVQA~\cite{akula2021crossvqa} presents a collection of test splits for assessing VQA generalisation in a variety of visual domains spanning from photographs (e.g., VQA v2) to real-world images taken by visually impaired users (e.g., VizWiz~\cite{gurari2018vizwiz}).

\subsection{The importance of Dialogue for Language Learning}

We argue that Interactive Grounded Language Learning is a plausible paradigm for training embodied agents to understand and use Natural Language. However, few of the papers that we analysed try to solve a task involving a multi-turn conversation between two agents. Additionally, the analysed papers can be further categorised based on the degree of interaction that they implement.

\subsubsection{Scripted dialogue policy}
Dialogue can be \emph{scripted} such as in the work by \cite{patki2018language} and  \cite{antunes2017learn}. In this case, the dialogue policy follows a predefined script that allows the model to complete the reference task. A modification of this setup involves having dedicated dialogue managers \textit{to learn when to learn}~\cite{yu2017learning,thomason:icra19}. Specifically, the dialogue policy relies on specific intents to detect when the model should update its internal representation as a result of an error in classification.  

%The CLEVR-dialogue\cite{kottur2019clevr} is the first dialogue datasets proposed to study visual coreference resolution. Specifically, the dataset is based on synthetic images from the CLEVR dataset \cite{johnson2017clevr}. There are two agents that play the game: an Answerer (A-er) who can ``see" the image and has the complete scene graph, and a Questioner (Q-er), who does not ``see" the image and is trying to reconstruct the scene graph over rounds of dialogue. They use the proposed functional programs to automatically generate a dialogue involving questions about the image. 
\subsubsection{Learned dialogue policy}
A more relevant setup for this work is one where the agent is supposed to learn \emph{when} and \emph{what} to generate directly from the data. For instance, \cite{agarwal2019community} propose a collaborative game where two bots are present. One of them -- A-bot -- receives an image and a caption. The other -- the Q-bot -- receives only the caption. Given the caption, it has to generate the image after asking the other bot some questions about it. In this way, it can create a representation that is suitable for retrieving, ranking or generating that image. In this way, no human supervision is required and a reward can be defined in terms of how close to the target the generated representation is (based on Euclidean distance). To tackle the \guesswhat{} dataset (see Section~\ref{sec:interactive_tasks}), \cite{de2017guesswhat} propose to train separate models for the 3 different tasks that need to be solved namely question generation, target object guessing and answer generation. \cite{shekhar2019beyond} propose an end-to-end neural model that uses a multi-task learning approach to jointly train the question generation and the guesser module. However, their model treats the image as a single feature vector generated by a pretrained CNN ignoring more fine-grained information present in real-world scenes. 

An approach that tries to more directly integrate the image features in the process of discovering the target object was presented in \cite{deng2018visual}. They propose a model that implements an iterative attention process that tries to co-ground the query feature to object-centric features associated with the image. However, they take into account only the guessing problem reducing its applicability to the full set of tasks that has to be solved in \guesswhat. A similar setup is also described in visual dialogue (e.g., \cite{kottur2019clevr,abhishek2017visdial}). For instance, CLEVR-Dialog~\cite{kottur2019clevr} has been defined to study visual coreference resolution in a multi-turn dialogue involving two agents. However, their main interest is to define models that are able to correctly track the state of the conversation rather than \emph{actually generating} the questions/answers.

\subsubsection{Multi-turn interaction}

Another important element is the relevance of multi-turn interaction and what is the granularity of the learning task. Specifically, some papers consider only single-turn interactions while others assume that the interaction can last up to $N$ steps.

Papers studying Natural Language evolution~(e.g., \cite{lazaridou2020multi,lazaridou2016multiagent,choi2018compositional,havrylov2017emergence}) can be considered as instances of a simplified dialogue namely a signalling game~\cite{lewis1969convention}. Here, they limit the interaction to single-turn exchanges where an agent (Speaker) generates a message and the other (Hearer) has to select the target image among a set of distractors. For instance, \cite{lazaridou2018emergence} studies the communication protocol developed by agents equipped with different types of visual encoders. In these studies, they are interested in the type of language developed by the agents as a result of the task completion objective. 

\begin{figure}[ht]
    \centering
    \includegraphics[scale=0.35, keepaspectratio]{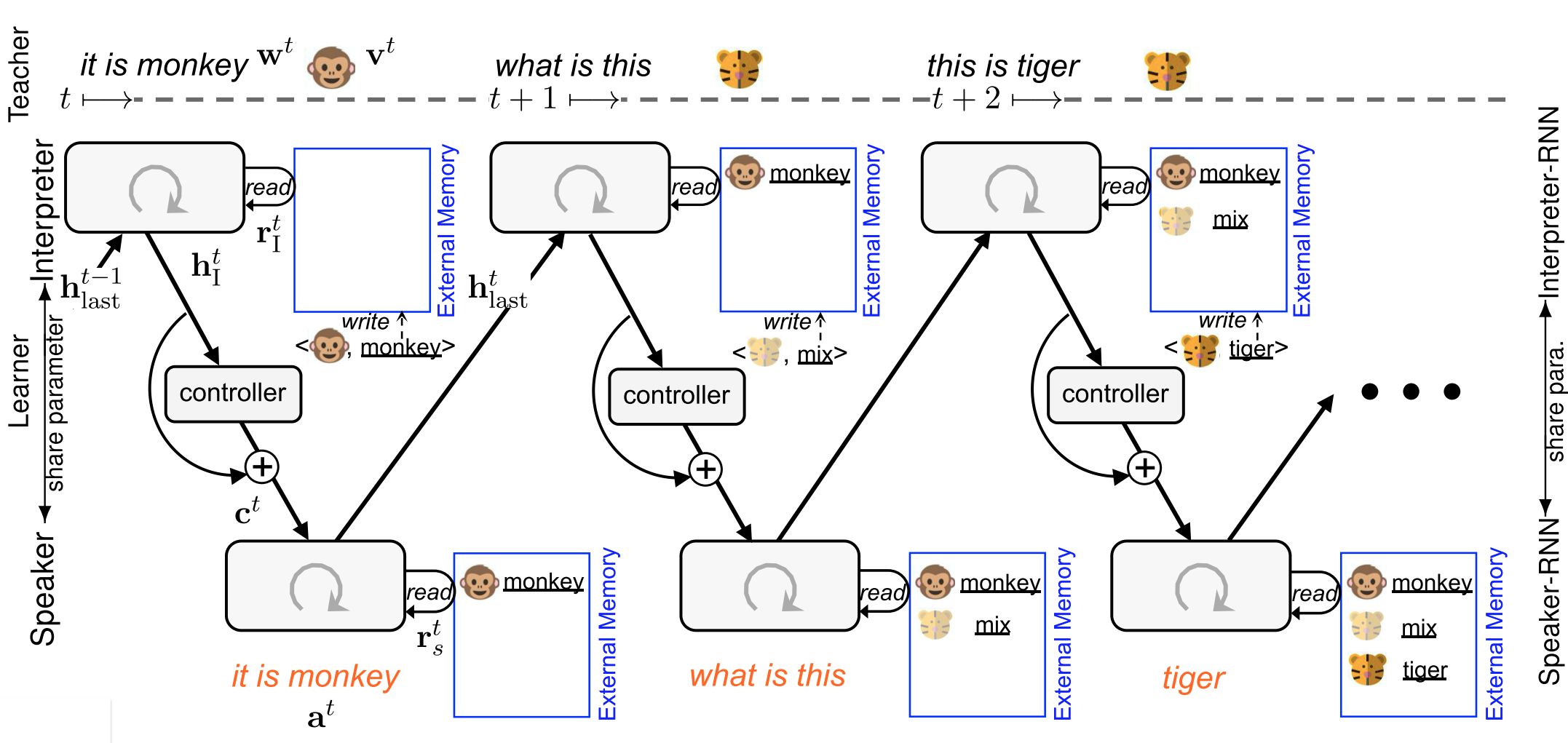}
    \caption{Model presented by \protect\cite{zhang2018oneshot}. The Speaker learns to mimic the Teacher by generating a statement associated with the images available. The Speaker has to learn animal concepts (and fruits) in a one-shot learning regime. After a single example of the class is shown to the Speaker by the Teacher, they are asked to classify an image associated with the same concept.}
    \label{fig:zhang_oneshot}
\end{figure}

For instance, a system that tries to learn word meanings in an interactive fashion was presented by \cite{zhang2018oneshot}. They present a synthetic task where an agent needs to learn the mapping between an image of an object, and a phrase, as shown in Figure \ref{fig:zhang_oneshot}. The images depict small-sized animals. They trained the model in a one-shot learning regime where the model learns to classify a test instance by relying on a single training instance of that class. To implement their model, they rely on a multimodal variant of the Neural Turing Machine read-write memory mechanism \cite{graves2014neural}. Specifically, it learns to associate a specific word (e.g., ``cat") with a corresponding visual representation (e.g., ``cat image"). In this paper, the dialogue involves learning multiple object categories at the same time. A single turn is required to learn the association between an image and a given category. Furthermore, the language used was synthetically generated by a grammar and therefore lacks the diversity of Natural Language. Despite these limitations, they tackle an important problem in the Machine Learning community namely \textit{Few-shot Learning} \cite{wang2019fewshot}: the agent needs to learn to maximise the performance P for a task T using a limited amount of experience E about it. In addition, they use a training strategy for their agent that combines both Imitation Learning and Reinforcement Learning. In particular, they assume that early in the training the agent is incapable of generating meaningful responses so it imitates what the teacher says. Later on in the training, it starts generating more interesting responses so they use Reinforcement Learning to provide a more precise signal about the quality of the generated response. This will give the agent the chance to learn to ``babble" and to correctly solve the conversational game. 

A model able to support a conversation composed of multiple turns can automatically generate extra training data from a very small bootstrap set. For instance, the work presented by \cite{misra2018learning} describes a \emph{learning by asking} paradigm where the model improves its VQA accuracy by learning to ask relevant questions about a given image. In this way, the model is able to retrieve data points that go beyond the training set and therefore learn better representations for the downstream task.

Depending on the task, it is debatable whether dialogue history plays an important role in the decision-making strategy of the learning agent. Particularly, tasks requiring single-turn interaction do not model the history in the first place. Additionally, even tasks based on multi-turn interaction do not necessarily require dialogue history. For instance, VisDial~\cite{abhishek2017visdial} is designed as a dialogue task but the questions that the humans develop can be answered by simply looking at the image. \cite{agarwal2020history} show that models having sophisticated history encoding mechanisms bring very marginal improvements due to the fact that only a portion of the data \emph{actually} requires the dialogue context. Therefore, special attention should be paid to the task design to make sure that dialogue history actually matters. 

\subsection{Embodiment \& Grounding}

\begin{figure}[h]
    \centering
    \includegraphics[scale=0.35,keepaspectratio]{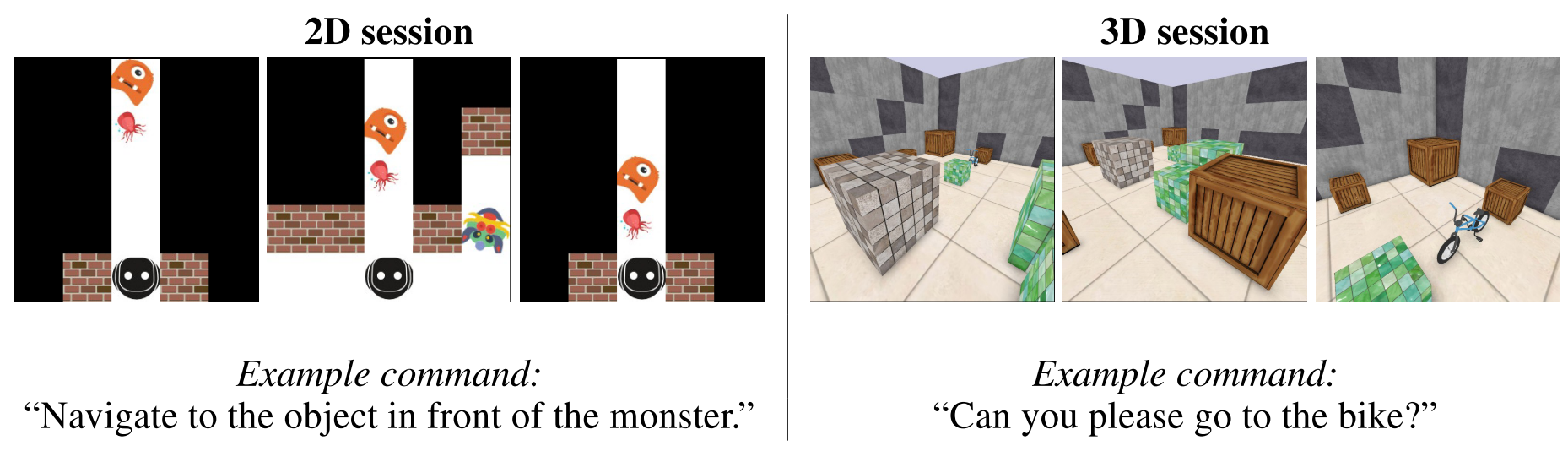}
    \caption{Examples of navigation tasks solved in the work presented by \protect\cite{yu2018guided}.}
    \label{fig:baidu_navtasks}
\end{figure}

In the analysed papers a common trend emerges. In most of the studies, the reference for grounding is an image that represents the context of a \emph{situated} interaction. As discussed in the previous section about action execution tasks (see Section~\ref{sec:interactive_tasks}), 3D simulators and virtual environments have been used to study the problem of grounding in visually rich contexts. \cite{yu2018guided} propose \textit{Guided Feature Transformation} which is a grounded language model extending classical CNN with an end-to-end differentiable component able to model the interactions between language and vision. They compute a recursive relationship where the input representation generated by the CNN is refined after J steps according to the words in the input command. They design several navigation tasks, as shown in Figure \ref{fig:baidu_navtasks}, in a 2D/3D environment: 1) \textit{nav}: navigate to a specific object; 2) \textit{nav\_nr}: navigate to an object near the specified one; 3) \textit{nav\_bw}: navigate to a location between two objects; 4) \textit{nav\_avoid}: navigate to any object but the specified one; and 5) \textit{nav\_dir}: navigate to an object specified by a relative direction w.r.t.\ another object.

\cite{chevalier2018babyai} propose a first model for the BabyAI environment. They use a GRU module to model the language instruction and a FiLM \cite{perez2018film} for the vision module which learns to condition the visual representation on the question representation. The agent accuracy on these tasks seems to be very high (nearly 100\%) posing some doubts related to the complexity of the world and of the language expressions used. Instead, \cite{bahdanau2018learning} describes a learning framework for instruction-conditional RL agents having to execute instructions in a grid world. In their RL setup, the agent learns from a reward that is derived from a reward model trained from expert demonstrations. 
% Removed as suggested by Yannis -- too many details about the model training
% Several instructions are supported: 1) move to precise location; 2) create a specific shape (by moving the blocks available in the grid); 3) create a specific shape with a given colour (when there are other blocks of another colours in the grid). The reward model parameters are learned with the model parameters using an adversarial regime. The reward model acts as a discriminator for defining goal and non-goal states. The predicted reward is considered 1 the prediction of the reward model is greater than 0.5, 0 otherwise. The agent is trained using \textit{AC3} \cite{mnih2016asynchronous} by optimising the expected cumulative reward provided by the trained reward model. The trained reward model is optimised using SL by considering a dynamic dataset composed of pairs (instructions, goal state) collected by letting the policy execute action in the environment. 
The proposed model is able to learn a reward model that performs well from a limited amount of data. Specifically, with a training set of only 8000 examples, the agent could reach a performance of 60\% that can be improved to reach optimal performance when a set of 100,000 examples is used to train the model.

Motivated by the desire to create artificial agents that can follow instructions in domestic environments, there have also been several efforts transitioning from grid-world setups to benchmarks for language-guided navigation in indoor 3D environments with photo-realistic images. Several models have been proposed in the literature to solve this problem. For instance, several Transformer-based models have been designed for this task (e.g., \cite{majumdar:eccv20,crossmap,cyclevlnbert}). An important ingredient for all of these models is a careful pretraining design. Specifically, to deal with the complexity of the language instructions, language-only pretrained weights are used (e.g., BERT~\cite{devlin2019bert}). Then, large-scale image-text pairs from the Web are leveraged to train the model. Using such additional pretraining data helps finetuning the cross-modal encoder that will be used for the downstream task evaluation in a photo-realistic environment such as Matterport3D~\cite{chang2017matterport3d}. 

Several projects have tried to investigate visual question-answering capabilities in 3D environments. For instance, the work presented by \cite{gordon2018iqa} introduces the IQA dataset that requires the agent to move in the environment in order to answer the questions. It extends VQA in several ways: 1) navigation in the environment; 2) understanding of the environment (object, actions, affordances); and 3) executing actions conditioned on the question. The main question types are 1) counting; 2) existence; and 3) spatial relationships. They present \textit{Hierarchical Interactive Memory Network (HIMN)}, a Deep Learning model that incorporates a hierarchy of modules: a \textit{Planner} intended as the decision-making component responsible for activating several \textit{task-specific modules}. The designed sub-modules are 1) navigator; 2) scanner; 3) detector; 4) manipulator; and 5) answerer. 
% The model can be trained jointly in an end-to-end fashion. However, they pre-train each component because each task can be considered as independent. Generalisation tests: Unseen environments tests the agent with questions that occur in 5 never-before-seen rooms, whereas the seen environments use the 25 training rooms but never-before-seen object placements and corresponding questions.
HIMN is unable to differentiate between an object being inside a container or on top of the container. From the language perspective, the questions are scripted and the language is relatively simple, a downside whose solution is considered as future work by the authors.

\begin{figure}
    \centering
    \includegraphics[scale=0.3]{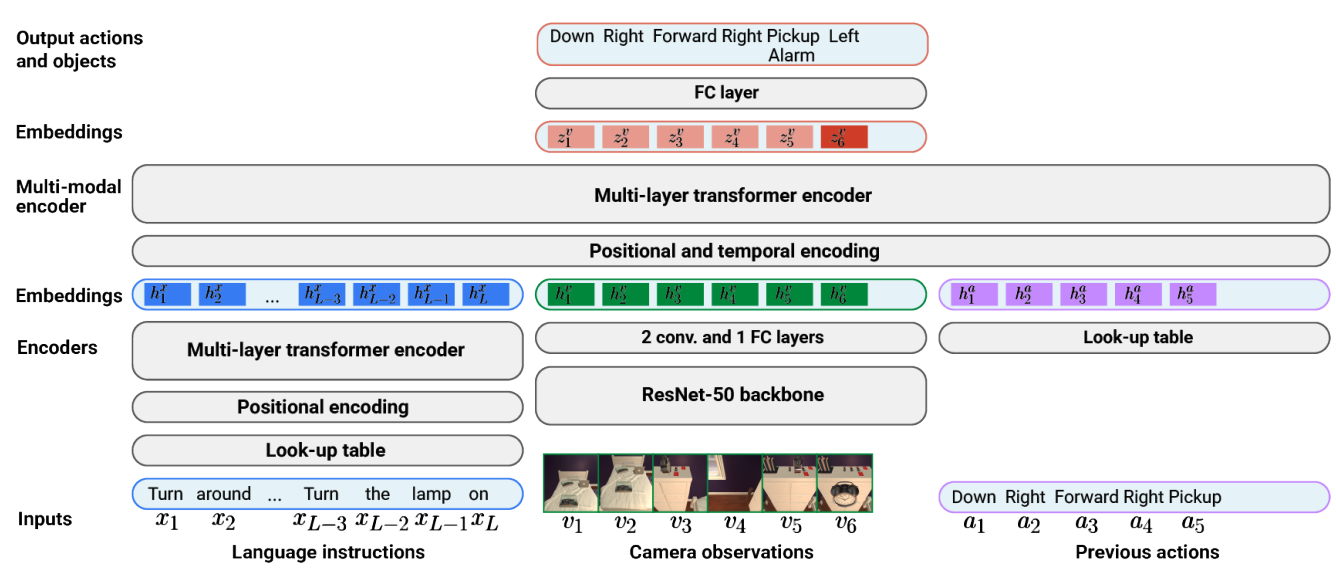}
    \caption{Architecture of the Episodic Transformer model presented by \protect\cite{et}. It jointly encodes language instruction, camera observations and previous actions using a dedicated multimodal encoder. The hidden state representation of the last state is then used to predict the action and the object class. Figure adapted from \protect\cite{et}.}
    \label{fig:et}
\end{figure}

With the aim of solving the ALFRED benchmark~\cite{alfred}, several models have been proposed. The proposed solutions are very specific to the task at hand. In the work presented by \cite{et}, they propose a Transformer-based architecture that encodes the CNN representation associated with each timestep, together with the language embeddings as well as the previous actions (see Figure~\ref{fig:et}). In this way, they can use the hidden state representations to generate actions conditioned on the context. Specifically, they use the hidden state of the last trajectory timestep, to predict the action to be executed as well as a possible object class that the agent should manipulate. The limitation of such an approach is that their architecture is specifically designed and trained only for ALFRED. For instance, for manipulation actions, they assume that only a single instance of a given object class is available in the manipulation area (e.g., the counter will contain only one ``apple" so no ambiguity is present). Therefore, it is possible to use an external object detector to detect the objects and then just select the one that we want to manipulate based on the class predicted by the model. This forces the model to reason only in terms of which actions can manipulate certain classes but it doesn't allow the model to have a good understanding of the actual objects that the agent has in its front view because its visual representation is not fine-grained enough. A more generic approach was proposed by \cite{suglia2021embodied} who designed a multi-modal Transformer able to model the fine-grained, and multi-modal ALFRED trajectories represented in terms of detected objects in the visual scene. As highlighted by \cite{shane-amazon}, navigating towards the target object is the hardest task in ALFRED because it can easily put the agent in a situation that the agent cannot recover from. Additionally, the navigation problem is exacerbated by the training regime used to train such models. Their policy is trained using supervised learning, mimicking human trajectories. However, at training time, agents do not have the ability to explore the environment to gain a better understanding of the objects and their location in the environment which might be important to successfully solve the tasks. 

\begin{figure}
    \centering
    \includegraphics[scale=0.25]{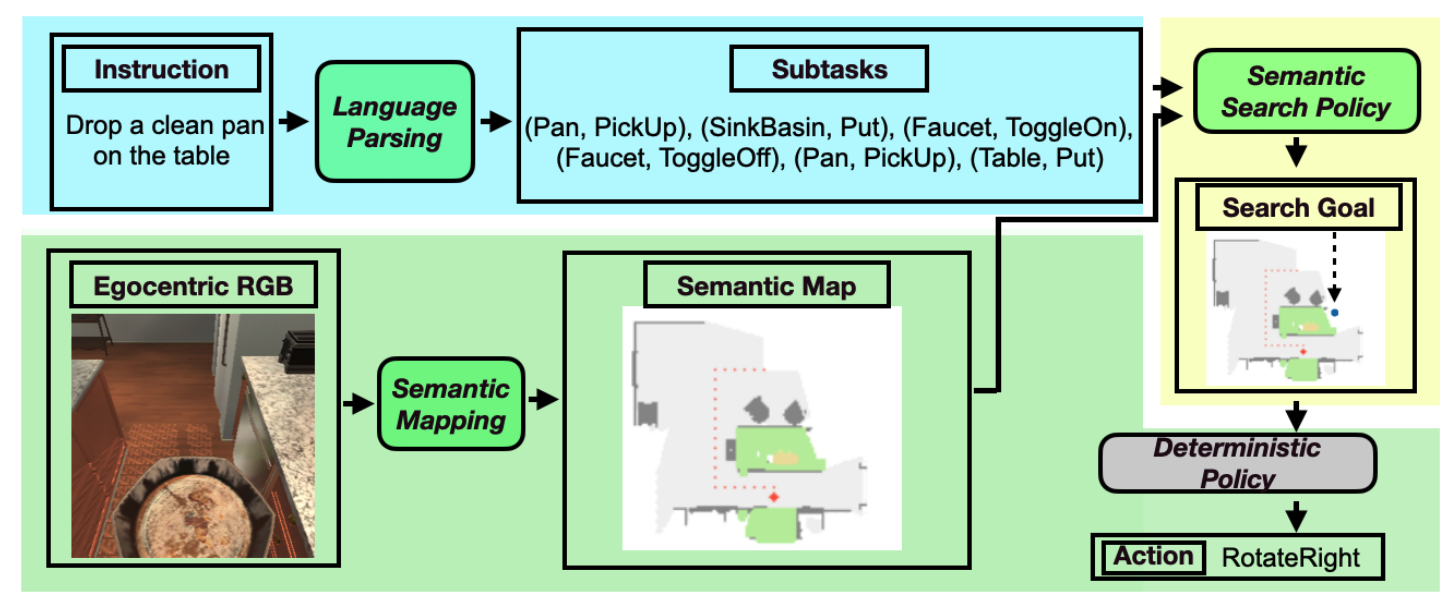}
    \caption{Overview of the FILM architecture for language-guided task completion in ALFRED. They developed a modular architecture composed of several pretrained components that are connected together to solve specific tasks. The goal instruction is the only linguistic information that is available. From it, an execution plan is derived where a semantic search policy looks for the relevant objects in the scene, while a deterministic policy actually executes the relevant actions. Figure adapted from \protect\cite{min2021film}.}
    \label{fig:film_alfred}
\end{figure}

Another possible solution to this problem is to avoid solving the problem in an end-to-end fashion, and instead divide it into multiple modules. For instance, as shown in Figure~\ref{fig:film_alfred}, modular architectures such as FILM~\cite{min2021film} can achieve better performance for this task. First, a dedicated module derives a semantic representation from the goal instruction. This will represent a high-level plan that the agent will have to execute. Specific navigation modules, paired with a map of the environment, are then used to navigate towards target objects. Despite its effectiveness compared to concurrent approaches~\footnote{FILM was 2nd in the ALFRED leaderboard as of 5th May 2022.}, its utility is debatable for models that are able to learn grounded representations that are useful for multiple tasks as well as domains. Indeed, such a model relies on specific algorithms that are used at inference time to complete the task. At the same time, it is important to understand that future models should have a way to derive a map of the environment in which they are located if they want to solve increasingly complex tasks with complete autonomy. 

\section{Discussion} \label{sec:discussion}

% Remember the original research questions that have to be referenced here!
% \begin{itemize}
%     \item How can Language Games be used for Visually Grounded Representation Learning?
%     \item What type of representations can be learned via Language games?
%     \item Are such representations generic enough to be transferred to other tasks?
%     \item What representations are useful for action-driven and event-driven language games?
% \end{itemize}
In this section, we provide a summary of our study of the literature on visually grounded language learning. As a result of our analysis, we derive several research questions that will be detailed in the following subsections. To answer these questions, novel evaluation benchmarks that are more suitable for grounded language learning as well as novel modelling solutions will be required. This survey represents an important reference for studying how to extend both models and related benchmarks to push the boundaries of grounded language learning. Finally, we compare our analysis with previous surveys in the field of V+L, and describe the importance of our categorisation to further study the field of V+L in the future.

\subsection{How does this survey relate to other V+L surveys?}

The main focus of this survey is to provide a systematic categorisation of several tasks that have been explored in the V+L community. In particular, we provide a categorisation based on 3 broad categories namely \textit{discriminative games}, \textit{generative games}, and \textit{interactive games}. This aims to provide a higher-level perspective of the different tasks that we believe will help researchers in sharing knowledge and in generalising results that have been produced by a specific task but that could be applied to another one as well. For instance, \cite{ferraro2015survey} presents a survey of image captioning datasets that dates back only to 2015 which can be categorised in the \textit{generative games} category of our survey. The work from \cite{li2022vision} presents a very detailed analysis of recent Transformer-based architectures that focuses on \textit{discriminative} and \textit{generative games} only ignoring the importance of interactive and embodied games for grounded language learning. A recent survey presented by \cite{gu2022vision} underlines the importance of embodiment and interactive learning. They analyse the field of Vision+Language navigation in great detail. However, as specified in our description of \textit{embodied interactive games}, visual navigation ignores an important part of learning which is object interaction. \cite{mogadala2021trends} presents a very broad survey of general vision and language research up to the year 2019, whereas this paper provides  a more up-to-date survey and analysis focused on Interactive Grounded Language Learning tasks. We consider Interactive Grounded Language Learning a new avenue for multimodal machine learning research. In particular, we argue for the importance of bringing together two relevant problems such as ``symbol grounding" and ``conversational grounding", which we believe should be studied together to build flexible learning agents that can smoothly adapt to different speakers and domains using Natural Language. 

Finally, we discuss the relationship between our survey and surveys which provide a broader view of artificial agents able to learn language. For instance, \cite{schlangen2019language} uses the notion of language games for describing Natural Language Processing tasks. This survey identifies the task of ``unrestricted situated language interaction" as ``a natural supremum" in NLP tasks which is akin to the \textit{interactive} language games that we consider in this work. In our survey, as first defined in \cite{bisk2020experience}, we underline the importance of considering NLP tasks that require that the agent can ``experience" the result of its actions in the world as well as communicate and collaborate with other agents uncovering the social role of Natural Language. 

\subsection{How can language games be used for Visually Grounded Representation Learning?}

In this survey, we have analysed several tasks as well as associated models that have been proposed to solve them. Each of them has a specific focus of interest in modelling visual scenes as well as language. Depending on the way the output was generated, we divided them into three types: discriminative, generative, and interactive tasks. We argue that interactive tasks are the most important for future work because they are closer to the language games scenario that we have described. Among them, we consider visual guessing games an important task for learning grounded representations of objects. Objects in an environment, and their attributes, are the first step in learning grounded representations. Other interactive language games are embodied ones. Particularly, we consider language-conditioned action execution in a 3D world as the most relevant benchmark for learning other important elements of grounded representations, such as actions and events. In particular, it will be important to consider new modelling techniques that facilitate the development of agents that can negotiate and coordinate plans of actions in the world.

\subsection{What type of representations can be learned via language games?}

When investigating the quality of distributed representations, it is possible to define several desiderata for distributed representations~\cite{bengio2013:representation_learning}. This is an important problem, especially for symbol grounding where we want to learn representations that are compositions of elements that are in turn grounded in the real world~\cite{harnad1990symbol}. As previously stated, if we focus only on increasing the performance on the downstream task which is evaluated based on a held-out test set, we might have an opaque understanding of the real system capabilities. Most datasets and models assume this scenario where the model is never confronted with novel object categories. Its generalisation abilities are assessed only via novel instances of known categories. Therefore, we want to make sure to investigate language games that are complex enough to let the agent learn grounded representations. A representation can be considered grounded when it is a composition of other grounded elements. Therefore, to assess this ability of the model, we should consider tasks involving zero-shot learning scenarios where novel categories are involved. In this case, the only way for the agent to solve the task is to use what it has learned during training in novel ways, therefore demonstrating its ability to systematically generalise. 
To mitigate this issue, \cite{suglia-etal-2020-compguesswhat} describes a novel multi-task evaluation framework that extends the traditional evaluation framework for grounded language learning to give value to the quality of the learned representations. Thanks to this evaluation framework, the authors have shown several limitations of current multimodal models used for situated visual guessing games. Therefore, it is very important that future benchmarks address this issue by designing evaluation setups that assess the model generalisation capabilities in a more systematic way~\cite{linzen2020can}.

\subsection{Are learned multimodal representations generic enough to be transferred to other tasks?}

From the works described in Section~\ref{sec:gll_models}, a common trend in the V+L literature is to propose a model architecture having inductive biases that can be applied to a specific task only.
However, it is important to underline that another property of high-quality distributed representations is their suitability for other tasks different from the one they have been originally trained on. Concretely, this implies that language games should favour the emergence of important properties of the real world that can be captured by the model. If such knowledge is generic enough, the agent should be able to reuse it for solving other tasks as well. For instance, \cite{suglia2021empirical} argue that visual guessing games can be developed as a generic transfer learning procedure for grounded language learning. This stresses the idea that more general architectures able to solve multiple tasks are required~\cite{brown2020language}. 

When considering the generalisation power of the developed models, it is also important to study the ability of these multimodal models to be applied to scenarios where not all the modalities are involved. For instance, it is still not clear whether current instances of multimodal architectures are able to bring an improvement in text-only scenarios such as Question Answering, Natural Language Inference, and so on. Several recent works have tried to address this issue (e.g., Vokenization~\cite{tan2020vokenization}) but they are not generic enough to be easily transferred to other tasks because they rely on additional components that are not part of the same trained architecture. It is important that future work investigates \textit{truly} multimodal models that can rely on multimodal modalities to create richer semantic representations that can be reused in scenarios that are different from the one they were originally trained on.

\subsection{What representations are useful for action-driven and event-driven language games?}

Among the grounded language learning tasks in the literature, visual guessing games represent the first step in a curriculum for learning object categories and their attributes. However, objects are only one part of the spectrum required for understanding. As argued in \cite{mcclelland2020placing}, the main target of language is the understanding of situations~\cite{barwise1981situations}. Situations are highly relational constructs composed of entities and relationships between them. Quoting \cite{mcclelland2020placing}: ``To understand a situation is to construct a representation of it that captures aspects of the participating objects, their properties, relationships and interactions, and resulting outcomes". This resonates with the idea that even the understanding of nouns (i.e., entities) requires composing affordances of them~\cite{glenberg2002grounding}. Following the idea of grounded cognition~\cite{barsalou2008grounded}, embodied experiences of concepts and events are required for the development of meanings. Thanks to the development of 3D environments, it is now possible to study embodied tasks and their relationship with grounded cognition. 

We have analysed several interactive tasks that study the problem of completing a task guided by Natural Language. First, language-guided navigation was proposed as a benchmark for embodied AI. Equipping the agent only with navigation actions is limiting its ability to learn grounded meanings. Therefore, several other benchmarks have been proposed to study more complex task-completion tasks involving manipulating objects as well. These environments represent an important avenue for embodied AI. Particularly, a general embodied AI architecture should be able to encode arbitrary long action sequences and make decisions upon them. However, simulated environments are limited by the capabilities of the physics engine, the object repertoire and the number of scenes available. Another interesting perspective for a learning agent is represented by videos in the third-person view (i.e., exocentric view). Learning from such multimodal inputs can expose the learning agents to many more situations that are useful in real-world contexts. 

\subsection{How does Symbol Grounding relate to Conversational Grounding?} \label{sec:symbol_conv_grounding}

As we highlighted in the introduction, interaction with the world and with other agents are key ingredients for language learning. When playing language games, speakers coordinate among themselves to refer to things in the world and achieve goals. Additionally, the ability to engage in conversation allows agents to be more robust in situations of uncertainty. Many recent works have indicated the importance of \textit{conversational grounding}~\cite{Brennan.Clark96} as foundational for the future of conversational systems. More importantly, many recent position papers have convincingly argued the importance of connecting symbols and conversational grounding to create truly collaborative systems~\cite{liu2013modeling,chai2016collaborative,schlangen2019grounded,benotti-blackburn-2021-grounding,chandu-etal-2021-grounding-grounding}. It is therefore important to define new tasks where symbol grounding and conversational grounding together have the role of first-class citizens, and where communicating agents have divergent information which they need to coordinate \cite{lemon-emecomm-iclr-2022}. The need for divergence is essential because it is what motivates agents to communicate to achieve specific goals. Additionally, this idea of divergence is related to the concept of Theory of Mind~\cite{astington2005language} which is used in this context to refer to social understanding. Following \cite{tomasello2010origins}, social understanding interacts with language communication in two ways depending on the type of social understanding desired. In particular, understanding the goals and perceptions of an agent will engage different language activities than the ones demonstrated when trying to understand one's desires and beliefs. 

Studying the interplay between symbol and conversational grounding in a systematic way will require dedicated data collection processes aimed at collecting data in realistic embodied environments from which it is possible to elicit both symbol and communicative grounding phenomena that a model can learn from. These phenomena require additional methods to model both the conversational context as well as to derive relevant signals from the language feedback provided by other agents. It is important to consider a learning scenario that is closer to a real-world scenario involving tasks providing an ecologically valid scenario~\cite{de2020towards} for studying multimodal conversational grounding. 

\section{Future Research Directions}

In this section, we will describe potential interesting research directions that we argue are worth exploring as a result of this  literature review. These research directions are important for the development of truly generalist agents that can \textit{perceive}, \textit{reason}, and \textit{act} in novel scenarios.

\subsection{Perceptual symbols}

Language is a means of communication, which has evolved by transmitting symbols grounded in perceptual experience (e.g., gestures, sounds, etc.). However, as described in Section~\ref{sec:gll_models}, current AI models learn a language only from symbolic representations derived from a tokenizer that are: 1) specific for a given language and are costly to transfer to low-resource languages, so limiting people's to access technology; 2) sensitive to noise (e.g., spelling mistakes); 3) hand-crafted because they do not represent language input in a multimodal way, as humans do.   

Thanks to recent innovations in Computer Vision, researchers have attempted to render language data as images, and use new Deep Learning architectures for Computer Vision to attempt this task. This solution was first proposed for Machine Translation by \cite{salesky-etal-2021-robust} who use a CNN architecture to derive visual representations of the input text that was rendered as an image. From these representations, the authors were able to translate to a target language without the need for an input tokenizer. \cite{rust2022language} generalised this approach by proposing \textit{PIXEL}, the first language model pre-trained with rendered images only. This model is pretrained using a self-supervised reconstruction loss that uses the image patches to learn latent representations of the language. In their paper, the authors demonstrate comparable performance with BERT on several tasks such as POS tagging and Natural Language inference. Most importantly, they show superior performance for very low-resource languages as well as languages that do not have a Latin script. 

Deriving language representations from the audio signal is another recent trend that goes under the name of \textit{textless NLP}~\cite{kharitonov2022textless}.  For instance, the work presented in \cite{harwath2017learning} demonstrates how to derive word-like representation units directly from visual and audio features. Additionally, there is a very interesting line of work interested in learning representations from vision, language, and audio data (e.g., \cite{zellers2022merlot}). However, this approach is still constrained by the use of a tokenizer which limits applicability to specific language scripts.  

Deriving language representations from multimodal inputs is essential to achieve more realistic NLP systems that are more widely applicable such as Spoken Language Understanding systems. Second,   spoken language carries many nuances (prosody, irony, anger, etc.) and expressive vocalisations (laughter, yawning, etc) that are not captured by text. Additionally, modelling language via the visual modality can allow the implementation of more sophisticated agents that are able to understand the content of websites, forms, and other media that contain multimodal content. In general, modelling language directly from other perceptual modalities has the potential to make AI applications more natural and expressive. To advance the state of the art in Multimodal Representation Learning, we should aim to design \textit{perceptual symbol systems} whose aim is to derive language representations from multimodal perceptual experiences in a process akin to the seminal work from \cite{barsalou1999perceptual}. 

\subsection{Multimodal robustness}

To date, many large-scale V+L pretrained models have been proposed, and they are typically only used for V+L tasks. However, there are very few works (e.g., \cite{tan2020vokenization,singh2022flava}) that demonstrate the ability of such models to bring a performance gain when it comes to single-modality tasks (e.g., vision-only or language-only). Therefore, we believe that the new generation of V+L models should be able to effectively encode the input modalities as well as be robust to \textit{perturbations} of them (i.e., cases where such are potentially missing or are noisy). For instance, the work by \cite{ma2022multimodal} shows how multimodal models receiving noisy or missing modalities achieve performance that is inferior or just on par with unimodal models. Therefore, another important ingredient for the implementation of a sound and effective \textit{perceptual symbol systems} is the ability to encode and \emph{reconstruct} modalities when these are missing or noisy.

Crucially, it is important to investigate the robustness of multimodal representations in general. Much recent literature has tackled this problem from the point of view of domain generalisation (e.g., \cite{akula2021crossvqa}) as well as the effect of spurious correlations and noise (e.g., \cite{tu2020empirical}). This calls for a more systematic empirical investigation of the effect of such shifts in distribution and how they affect models across multiple tasks. For this reason, it is important to develop experimental frameworks that diagnose what matters in learning representations for V+L tasks (e.g., \cite{zhu-etal-2022-diagnosing,akula-etal-2020-words}).

\subsection{Interaction with the world and with other agents}

As we described in Section~\ref{sec:symbol_conv_grounding}, future V+L agents need to develop the ability to interact with the world around them. This is important because the ultimate goal is to create AI systems that can communicate with humans in Natural Language and develop a seamless symbiotic relationship with them. The current wave of Generative AI models has demonstrated that having conversational skills is essential for user-facing systems such as ChatGPT \cite{OpenAI2022IntroducingChatGPT}. 

In this scenario, we will assume that the agent is embodied in an environment (either simulated or realistic) and that there are one or multiple agents in it. The agent, motivated by a goal $g$, will have to produce a sequence of actions that will achieve their goal. There could be two types of actions that the agent can perform: \textit{physical actions} and \textit{verbal actions}. Physical actions involve a change in the agent's position or a change in the environment around them. In both cases, the agent will be able to \textit{observe} the results of their actions and derive signals from these changes that can be used for learning purposes. Verbal actions instead are intended as utterances that are formulated when engaging in conversations with other agents. This is required to acquire information and resolve possible ambiguities arising in the process of conversational grounding. Additionally, we also envision other types of verbal actions that involve \textit{inner monologue}~\cite{vygotsky2012thought}. This form of \textit{self-referential dialogue} could be very effective if intended as a form of planning that allows the agent to foresee the outcome of certain actions, and recover from possible problems. This approach has already been implemented in some form in text-only language models via \textit{chain of thought}~\cite{wei2022chain,yao2022react} as well as using text-only models for planning in 3D simulated environments (e.g., \cite{huang2022inner}).

\section{Conclusions} \label{sec:conclusions}

In this work, we surveyed datasets and tasks proposed to study the problem of visual symbol grounding in multimodal machine learning models. Following the language games proposed in the Talking Heads experiments~\cite{steels2015talking}, we proposed a categorisation of these tasks into 3 main categories: discriminative, generative, and interactive language games. We argue that \textit{interactive games} represent the most relevant class of language games to study the problem of grounded language learning. Additionally, we studied the properties and the skills required for solving the proposed language games. We find that language games situated in 3D environments enabling navigation and object manipulation represent an important step forward. In this way, artificial agents can learn object reference in the world (i.e., required when navigating in an environment), as well as the effects of certain actions on certain objects (i.e., observed when manipulating objects) which are essential for the development of rich grounded meaning representations.

After categorising all the available tasks, we analysed the multimodal machine learning models that have been proposed to solve them. We completed an analysis based on several key characteristics that these models have.  We find that in order to support interactive language games, we need more sophisticated ways of encoding rich multimodal context, as well as supporting conversations aimed at collecting important information for the task at hand. In addition, we highlighted the need for multimodal models able to generate representations that are meaningful even in scenarios where some modalities are missing (i.e., text-only models), or when some of the modalities are noisy. This is an essential ingredient for developing systems that are resilient to adverse scenarios that are very common in the real world.

As a result of this survey, we have identified several weaknesses primarily concerned with the datasets used to train such models. The development of sophisticated 3D engines has sparked research into embodied multimodal models able to solve tasks in simulated environments. This represents an important direction for future work with the aim of developing more ecologically valid studies~\cite{de2020towards} of how grounded meanings can be developed. Additionally, to adequately study language learning, we stressed the idea that we require datasets that require multi-agent communication skills. These are essential to account for recovery from possible breakdown and uncertainty during task solving. Moreover, it will facilitate agents to learn new tasks by learning from interaction~\cite{gluck2018interactive}. 

\section{Acknowledgements}

We would like to thank Arash Eshghi and Raquel Fernandez for their feedback on a preliminary version of this manuscript which was part of the first author's PhD thesis~\cite{suglia2022visually}.

\bibliography{survey}
\bibliographystyle{theapa}

\end{document}